\documentclass{article}

\usepackage{PRIMEarxiv}

\usepackage[utf8]{inputenc} 
\usepackage[T1]{fontenc}    
\usepackage{hyperref}       
\usepackage{url}            
\usepackage{booktabs}       
\usepackage{amsfonts}       
\usepackage{nicefrac}       
\usepackage{microtype}      
\usepackage{lipsum}
\usepackage{fancyhdr}       
\usepackage{graphicx}       
\graphicspath{{media/}}     
\usepackage{authblk}
\usepackage{graphicx,verbatim}
\usepackage{amsmath}    
\usepackage{bm}         
\usepackage{mathtools}
\usepackage{booktabs}   
\usepackage{graphicx}
\usepackage{hyperref}
\usepackage{algorithm}
\usepackage{algorithmic}
\usepackage{multirow}
\usepackage{tabularx}
\usepackage{array}
\usepackage{graphicx,booktabs,multirow,makecell}
\usepackage{subfigure} 
\usepackage{algorithmic}
\usepackage{array}
\usepackage{algorithm}
\usepackage{utfsym}
\usepackage{xcolor}
\usepackage{multirow}
\usepackage{graphicx,verbatim} 
\usepackage{amssymb}       
\usepackage{mathtools}
\usepackage{booktabs}   
\usepackage{graphicx}
\usepackage{hyperref}
\usepackage{algorithm}
\usepackage{algorithmic}
\usepackage{multirow}
\usepackage{tabularx}
\usepackage{colortbl}
\usepackage{array}
\usepackage{booktabs}
\usepackage{caption}

\title{Region-aware Spatiotemporal Modeling with Collaborative Domain Generalization for Cross-Subject EEG Emotion Recognition}

\author[1,$\dagger$]{Weiwei Wu}
\author[2,$\dagger$]{Yueyang Li}
\author[1,*]{Yuhu Shi}
\author[1]{Weiming Zeng}
\author[3]{Lang Qin}
\author[4]{Yang Yang}
\author[5]{Ke Zhou}
\author[6]{Zhiguo Zhang}
\author[2]{Wai Ting Siok}
\author[2,*]{Nizhuan Wang}

\affil[1]{Laboratory of Digital Image and Intelligent Computation, Shanghai Maritime University, Shanghai 201306, China}
\affil[2]{Department of Language Science and Technology, The Hong Kong Polytechnic University, Hung Hom, Kowloon, Hong Kong SAR, China}
\affil[3]{School of Chinese as a Second Language, and Center for MRI Research, Academy for Advanced Interdisciplinary Studies, Peking University, Beijing 100871, China}
\affil[4]{CAS Key Laboratory of Behavioral Science, Center for Brain Science and Learning Difficulties, Institute of Psychology, Chinese Academy of Sciences, Beijing 100101, China; and Department of Psychology, University of Chinese Academy of Sciences, Beijing 100049, China}
\affil[5]{Beijing Key Laboratory of Applied Experimental Psychology, Faculty of Psychology, Beijing Normal University, Beijing 100875, China}
\affil[6]{Institute of Computing and Intelligence, Harbin Institute of Technology Shenzhen, Shenzhen 518000, China}
\affil[$\dagger$]{Co-first authors}
\affil[*]{Correspondence: syhustb2011@163.com; wangnizhuan1120@gmail.com}

\begin{document}
\maketitle

\begin{abstract}
Cross-subject EEG-based emotion recognition (EER) remains challenging due to strong inter-subject variability, which induces substantial distribution shifts in EEG signals, as well as the high complexity of emotion-related neural representations in both spatial organization and temporal evolution. Existing approaches typically improve spatial modeling, temporal modeling, or generalization strategies in isolation, which limits their ability to align representations across subjects while capturing multi-scale dynamics and suppressing subject-specific bias within a unified framework. To address these gaps, we propose a Region-aware Spatiotemporal Modeling framework with Collaborative Domain Generalization (RSM-CoDG) for cross-subject EEG emotion recognition. RSM-CoDG incorporates neuroscience priors derived from functional brain region partitioning to construct region-level spatial representations, thereby improving cross-subject comparability. It also employs multi-scale temporal modeling to characterize the dynamic evolution of emotion-evoked neural activity. In addition, the framework employs a collaborative domain generalization strategy, incorporating multidimensional constraints to reduce subject-specific bias in a fully unseen target subject setting, which enhances the generalization to unknown individuals. Extensive experimental results on SEED series datasets demonstrate that RSM-CoDG consistently outperforms existing competing methods, providing an effective approach for improving robustness.
The source code is available at https://github.com/RyanLi-X/RSM-CoDG.
\end{abstract}

\section{Introduction}
Emotion is a core psychobiological construct pivotal to human adaptation. It underpins critical domains such as survival, cognition, well-being, and social interaction, constituting a complex dynamic process of integrated cognitive appraisal, affective experience, and physiological response to personally relevant events. Emotion recognition refers to the computational task of inferring an individual’s cognitive-affective state -- a complex integration of emotion, mood, and reactive psychology -- from multi-modal observations, including physiological signals, facial expressions, and behavioral patterns \cite{1,2,3}. However, many traditional approaches primarily rely on non-physiological cues such as facial expressions or vocal tones, which are subject to high variability, voluntary control, and subjective interpretation \cite{5}. Electroencephalogram (EEG) provides a direct, non-invasive measurement of brain activity with high temporal resolution \cite{li2024neural}, allowing for the characterization of rapid neural dynamics and a more objective approach to emotion analysis \cite{8,chen2025eeg,gong2025lerel}. Despite its advantages for emotion detection, the complexity of EEG signals and high inter-subject variability pose significant challenges for building models with strong generalization, especially in cross-subject applications.
\begin{figure}
\centering
\includegraphics[width=0.7\linewidth]{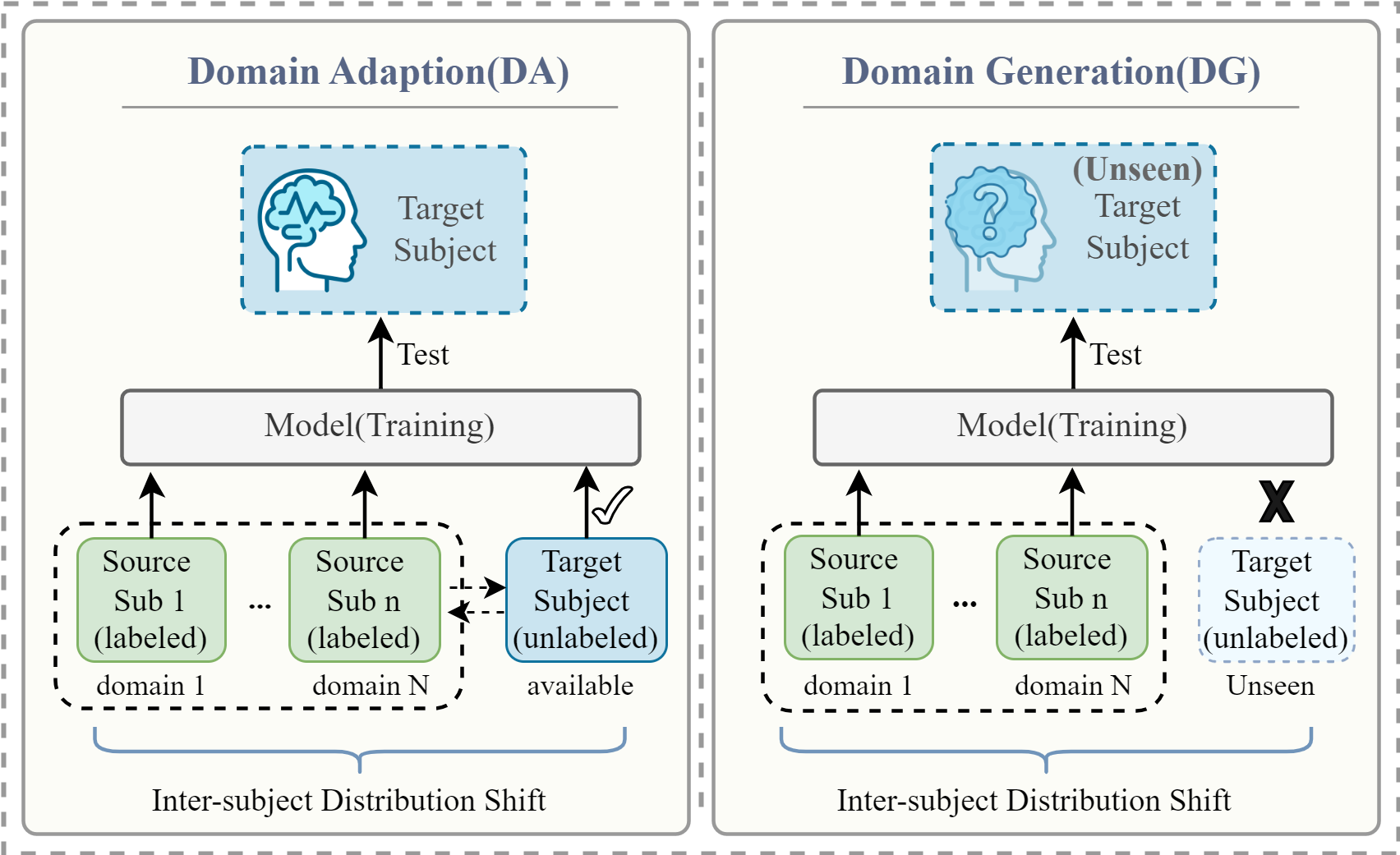}
\caption{Conceptual comparison of domain adaptation (DA) and domain generalization (DG) in cross-subject EEG emotion recognition. DA aligns source and target feature distributions using unlabeled target data during training, whereas DG learns a subject-invariant representation from multiple source subjects and generalizes directly to unseen targets without any target-domain data.}
\label{fig:figDG}
\end{figure}

In recent years, deep learning has driven significant progress in EEG-based emotion recognition. However, the systematic characterization of its spatiotemporal dynamics remains a critical bottleneck. For instance, spatial models that employ graph structures based on physical distance or signal correlation \cite{9,10,11} often treat the brain as a homogeneous network, neglecting the inherent regionalization and functional specialization underpinning emotional processing \cite{12,13,yu2021novel}. In addition, while neuroscientific research consistently demonstrates that emotional responses involve specialized and lateralized contributions from distinct brain regions \cite{14,15}, existing graph models often lack explicit neuroscientific priors needed to steer attention toward both intra-regional functional coordination and inter-regional specialized contributions. This omission ultimately limits their cross-subject generalization. Conversely, EEG signals are inherently dynamic, fluctuating with cognitive-emotional states over time. In this temporal dimension, emotional evolution is encoded and characterized by short-term continuity interspersed with occasional shifts driven by external stimuli or cognitive appraisal. This necessitates computational models capable of concurrently capturing both fine-grained local dynamics and long-range global dependencies \cite{18,19}. However, existing methods for temporal modeling often struggle to effectively integrate these complementary dynamic features. 

These bottlenecks in feature extraction amplify a major obstacle to practical application: inter-subject distributional differences. Data from different subjects exhibit significant distributional shifts, causing models trained on source domains to suffer from substantial performance degradation on unseen subjects \cite{wang2017novel}. As illustrated in Fig.~\ref{fig:figDG}, conventional domain adaptation (DA) methods attempt to mitigate these inter-subject shifts by leveraging target-domain data to align feature distributions between source and target domains. However, this reliance on target-subject data is often impractical in real-world EEG applications. In contrast, domain generalization (DG) aims to learn robust and transferable representations exclusively from multiple source domains, enabling models to generalize effectively to unseen target subjects without requiring any target-domain data during training. Numerous methods have emerged in this field, such as adversarial training for domain-invariant feature learning \cite{21,22}, explicit distribution matching via statistical alignment \cite{23}, and meta-learning or data augmentation to simulate domain shifts \cite{24,25,26}. Yet, most existing methods focus on optimizing a single dimension, lacking a unified mechanism to jointly and synergistically suppress subject-specific biases across multiple dimensions.

In response to these challenges, we propose a novel neuroscience-informed end-to-end architecture, RSM-CoDG. This framework is designed to mitigate subject-specific bias and extract transferable emotional neural representations by synergistically optimizing all objectives within a unified framework. In the spatial dimension, we propose the Region-aware Graph Representation Module (RGRM) to overcome the homogeneity of conventional GNNs. The RGRM systematically incorporates neuroscience priors from functional brain region partitioning through an intra-region collaborative–sparse dual-pathway mechanism, designed to capture complementary information processing patterns. Concurrently, to model emotional dynamics, we introduce the Multi-Scale Temporal Transformer (MSTT). It utilizes a parallel temporal encoding architecture: a local encoder employs masked attention to characterize the smooth evolution of emotional states, while a global encoder leverages periodic sparse attention to capture salient long-range dependencies triggered by external stimuli. To mitigate inter-subject shifts, our Collaboratively optimized Domain Generalization strategy (CoDG) transcends single-dimension alignment. It formulates a multi-objective regularization framework that jointly enforces distribution alignment, spatiotemporal attention consistency, and feature structure disentanglement. This approach learns more discriminative and transferable emotion representations without requiring access to target-domain data. The contributions of this paper can be summarized as follows:
\begin{itemize}
    \item This study introduces a unified, neuroscience-inspired modeling perspective, formulating a coherent framework that jointly addresses the complex spatiotemporal dynamics of emotion and inter-subject distribution shifts. This work establishes a novel theoretical foundation for learning cross-subject emotional neural representations.
    
    \item Informed by this perspective, we propose RSM-CoDG, an end-to-end framework that integrates region-aware spatial encoding, multi-scale temporal modeling, and collaborative domain generalization. This integration provides a systematic solution to the challenge of cross-subject EEG-based emotion recognition.

    \item Experimental results show that the proposed RSM-CoDG framework achieves state-of-the-art (SOTA) performance on the SEED, SEED-IV, and SEED-V benchmark datasets for emotion recognition.
    
\end{itemize}

The remainder of this paper is organized as follows. Section \ref{rw} reviews related work. Section \ref{pm} details the proposed RSM-CoDG framework. Section \ref{ex} describes the experimental setup and implementation details. Section \ref{raa} reports the experimental results and provides in-depth discussion supported by visualization and interpretability analyses. Section \ref{con} concludes the paper and outlines directions for future work.
\begin{figure*}[!t]
\centering
\includegraphics[width=\linewidth]{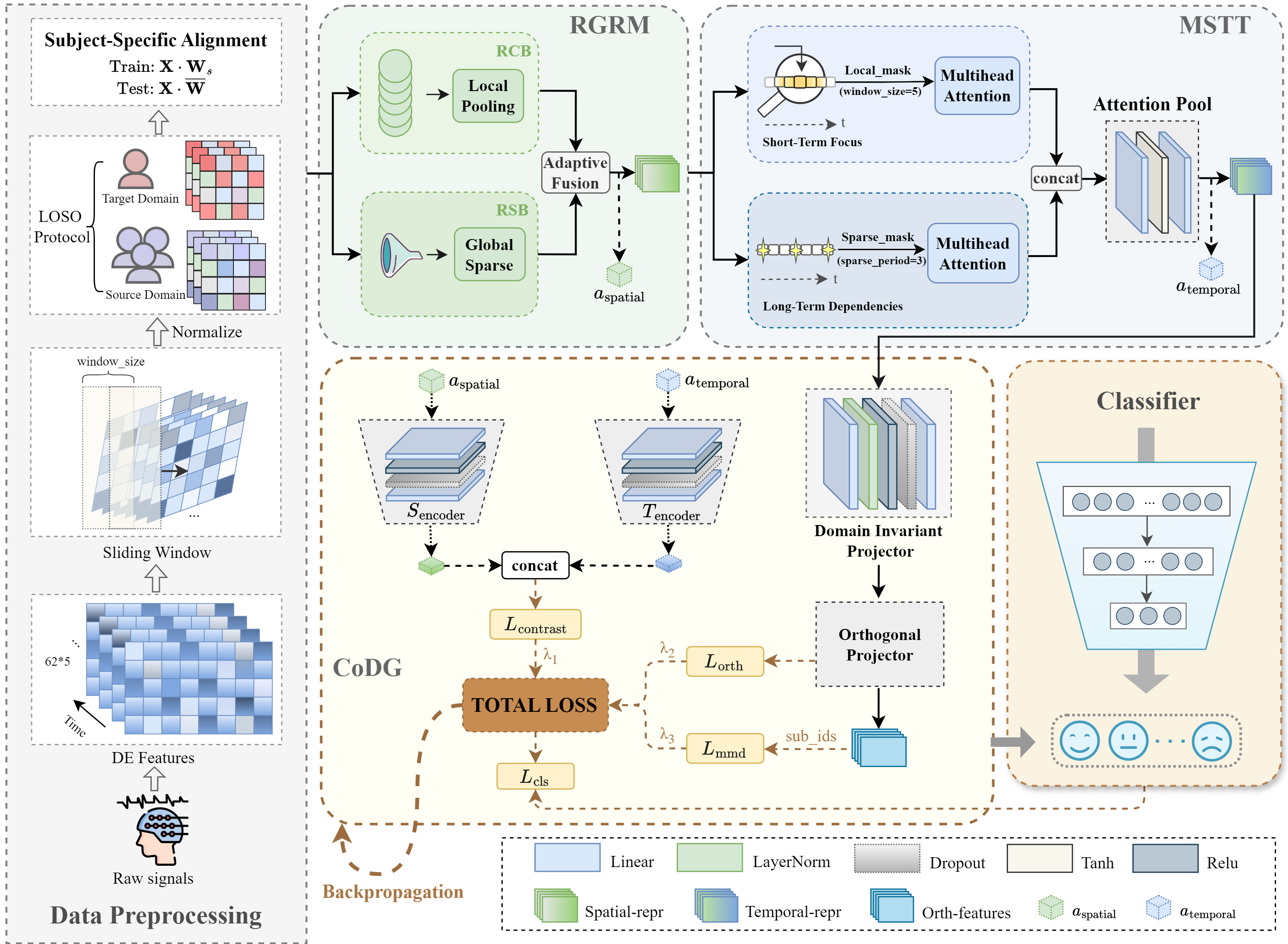}
\caption{Overall framework of the RSM-CoDG Model.}
\label{fig:fig1}
\end{figure*}

\section{Related Work}
\label{rw}
\subsection{Spatiotemporal Representation Learning in EEG Emotion Recognition}
The paradigm of EEG emotion recognition has evolved from handcrafted features to end-to-end representation learning, with a primary focus on modeling spatial dependencies and characterizing temporal dynamics. In the spatial dimension, graph-based approaches have emerged as a dominant paradigm. These methods treat electrodes as nodes and capture spatial dependencies in emotional brain activity by modeling connections between them. Early studies often relied on static topologies defined by physical distances or empirical functional connectivity, using Graph Convolutional Networks (GCNs) for spatial feature extraction \cite{11}. To address the limitations of fixed topologies in adapting to inter-subject variability and dynamic connectivity, subsequent research introduced dynamic graph neural networks that learn adaptive adjacency matrices based on feature similarity \cite{10}. More recent studies have further incorporated structural constraints, such as sparsity and symmetry, into learnable graphs to regularize their degrees of freedom \cite{29}. Notably, recent trends reflect a deeper focus on structured semantic modeling. For instance, the Region-to-Global Spatio-Temporal Neural Network (R2G-STNN) was proposed to model spatial relationships among EEG electrodes both within and across brain regions \cite{30}. Similarly, the Emotion Transformer (EmT) represents EEG as temporal graphs and integrates a multi-scale GCN with a contextual Transformer for joint spatial–temporal modeling \cite{31}. Despite these advances, existing approaches often fail to adequately model the specialized and differential contributions of distinct functional brain regions to emotional representation \cite{13}.

In terms of temporal modeling, the emotion elicitation process exhibits significant dynamics, with key emotional information embedded within the temporal evolution of EEG sequences \cite{32}. Early studies predominantly employed recurrent neural networks and their gated variants, such as LSTM and GRU, to model short-term temporal dependencies \cite{33,kang2025hypergraph}. 
In recent years, the Transformer architecture, with its strength in modeling global dependencies via self-attention, has emerged as a dominant paradigm \cite{37}. This has led to innovations such as the MEEG-Transformer, which mines complementary cross-domain information from temporal, frequency, and wavelet domains \cite{38}, and the EEG Conformer, which combines local convolutional operations with global self-attention to jointly capture short-term patterns and long-range contextual dependencies \cite{42}. Subsequently, dynamic-attention-based approaches have been proposed to explicitly model EEG state transitions by adaptively adjusting attention over successive time segments \cite{shen2025dynamic}. Nevertheless, emotional neural dynamics inherently involve both rapid transient variations and slower modulations, which single-scale attention mechanisms struggle to model effectively.

\subsection{DA and DG in EEG-based Emotion Recognition}
In cross-subject emotion recognition, significant domain shifts between individuals inevitably cause distribution discrepancies and performance degradation. This challenge has driven research into two primary transfer learning paradigms: DA and DG. DA methods mitigate these shifts by leveraging unlabeled target data to learn domain-invariant features, a strategy prominently exemplified by adversarial training. Representative approaches include the Two-level Domain Adversarial Neural Network (TDANN), which innovatively combines Maximum Mean Discrepancy with domain discriminators for global and local feature alignment \cite{43}, and DASC, which alleviates inter-subject discrepancies by clustering subjects with similar EEG emotion patterns and performing domain adaptation within each cluster \cite{liudomain}. 
For multi-source scenarios, the Multi-Source Marginal Distribution Adaptation (MS-MDA) method adopts a one-to-one alignment strategy between each source domain and the target domain to robustly handle domain-specific components \cite{23}.

However, the reliance of DA methods on target-domain data poses practical limitations for real-world deployment. This has driven a shift toward DG, which aims to achieve robustness to unseen subjects without requiring any target data. In this direction, meta-learning offers an effective paradigm. The MLDG framework, for example, simulates domain shifts via meta-train and meta-test splits to optimize generalization to virtual unknown domains \cite{24}. Moreover, self-supervised learning has demonstrated significant potential for extracting invariant features. The CLISA approach treats EEG samples from different subjects as positive pairs, explicitly maximizing their representation similarity through contrastive learning mechanisms \cite{46}. Similarly, the Denoising Mixed Mutual Reconstruction (DMMR) framework leverages data augmentation and mutual reconstruction principles, constructing hybrid subject features under a denoising objective to improve accuracy without additional parameters \cite{25}. Despite these advances, most existing DG strategies rely on a single constraint mechanism, such as adversarial, contrastive, or meta-learning objectives. Consequently, they fail to systematically mitigate the multi-factorial subject-specific biases inherent in EEG data.

\section{Method}
\label{pm}
\subsection{Overall Architecture}
In this study, we propose a novel cross-subject EEG emotion recognition framework, RSM-CoDG, whose overall architecture is illustrated in Fig.~\ref{fig:fig1}. The framework integrates three main components: (1) a Region-aware Graph Representation Module (RGRM), (2) a Multi-Scale Temporal Transformer (MSTT), and (3) a Collaboratively Optimized Domain Generalization Strategy (CoDG). The pipeline begins by applying a lightweight subject-specific alignment layer to the input EEG features for individualized linear calibration. The calibrated features are then fed into the RGRM for region-aware spatial representation learning. Next, the MSTT processes the spatial features to model multi-scale temporal dependencies. Throughout training, the CoDG strategy imposes collaborative regularization on the learned representations to enable robust cross-subject generalization.
\begin{figure*}
\centering
\includegraphics[width=\linewidth]{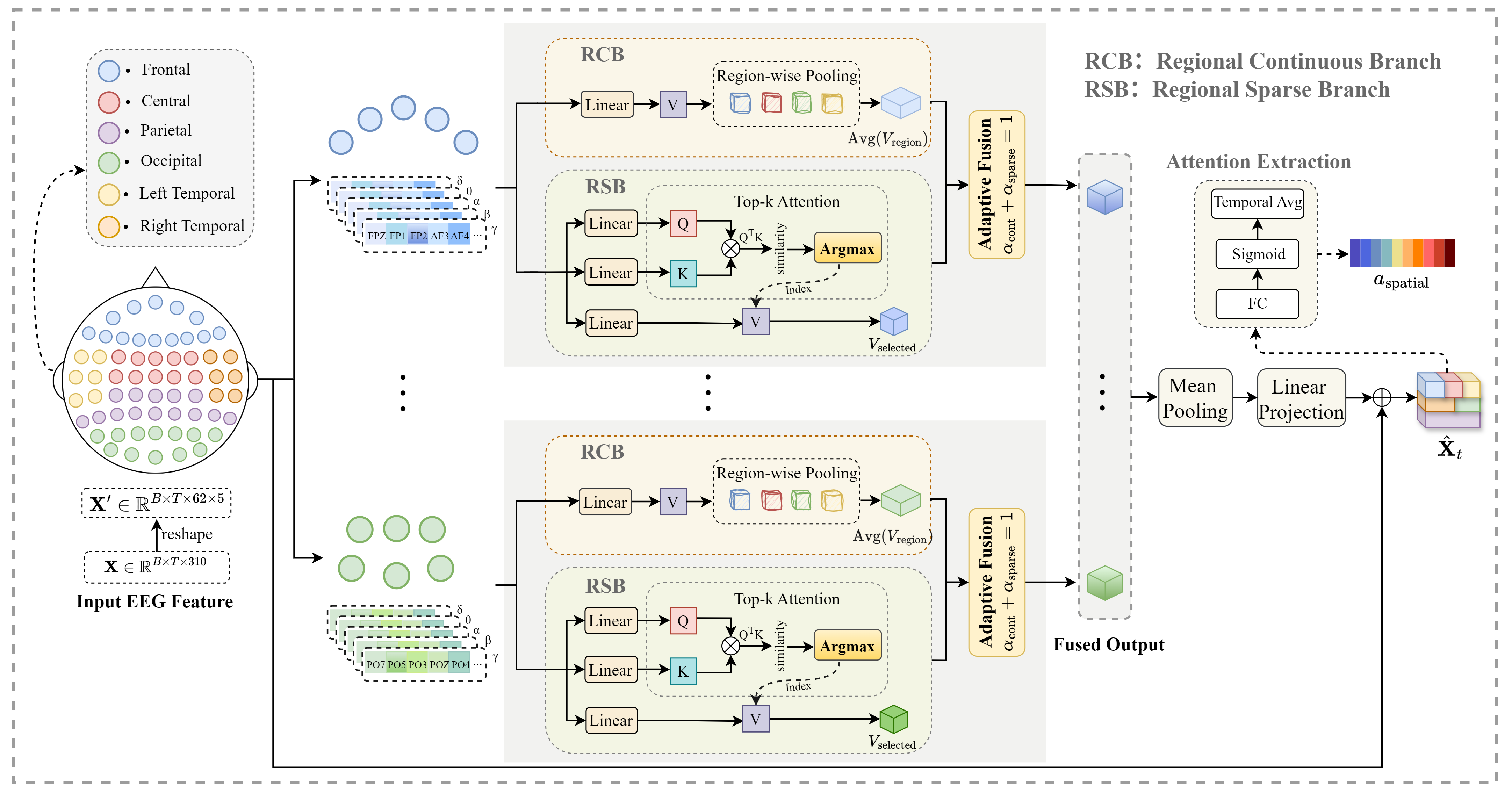}  
\caption{Network structure diagram of RGRM.}
\label{fig:fig2}
\end{figure*}

\subsection{Region-aware Graph Representation Module}
\begin{figure}
\centering
\includegraphics[width=0.5\linewidth]{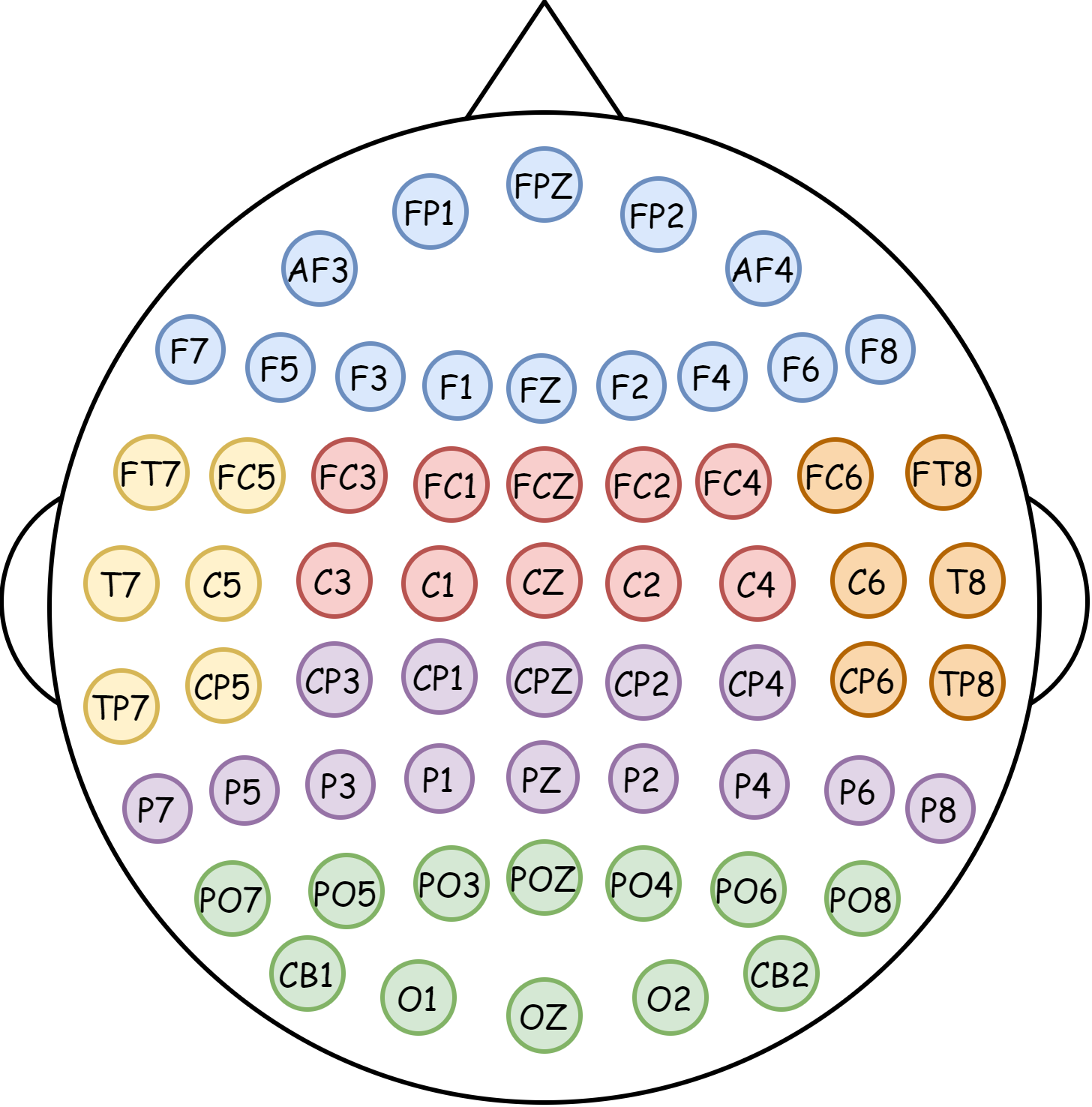}
\caption{Illustration of the 62 EEG electrodes partitioned into six functional brain regions. Electrodes sharing the same color belong to the same brain area.}
\label{fig:fig3}
\end{figure}

The RGRM is designed to incorporate semantic structural priors from functional brain region partitioning into spatial feature modeling. Using functional brain region divisions based on the international 10–20 system as a structural inductive bias, RGRM constructs a structured spatial attention mechanism that explicitly models region-specific interactions while preserving electrode-level signal integrity. This enables the module to jointly capture overall regional activation trends and discriminative responses of key local electrodes within functional brain areas. The detailed process is illustrated in Fig.~\ref{fig:fig2}. Specifically, guided by the international 10–20 system and brain functional zoning theory, we partition the 62 electrodes into six functional regions: frontal, central, parietal, left temporal, right temporal, and occipital, as shown in Fig. \ref{fig:fig3}. The electrode distribution and core functional description for each region are summarized in Table \ref{tab:table1}. The partitioning serves as a neuroscientific prior in region-aware spatial modeling, enhancing the physiological interpretability of features and their cross-subject consistency. The scheme comprehensively accounts for functional lateralization and specificity in emotional processing.

Prior to RGRM, we incorporate a lightweight subject-specific linear transformation layer to mitigate distributional shifts in EEG representations across different subjects. Specifically, for each subject $s$, we learn a dedicated linear transformation matrix $\mathbf{W}_\text{s} \in \mathbb{R}^{F \times F}$. The calibration operation for the input feature $\mathbf{X}_{\text{raw}}$ is defined as follows: During training, $\mathbf{X} = \mathbf{X}_\text{{raw}} \cdot \mathbf{W}_\text{s}$, during testing, $\mathbf{X} = \mathbf{X}_\text{{raw}} \cdot \overline{\mathbf{W}}$, where $\overline{\mathbf{W}} = \frac{1}{S}\sum_{s=1}^{S} \mathbf{W}_\text{s}$ and $S$ denotes the total number of subjects. This design enables the model to learn individualized linear mappings for each subject during training while employing an averaged transformation during inference to ensure generalization to unseen subjects.

\begin{table}[!t]
\centering
\footnotesize
\renewcommand{\arraystretch}{1.05}
\caption{Correspondence between functional brain regions and EEG electrodes based on neuroanatomical and functional zoning, with core region-level functions relevant to emotion-related cognitive and sensory processing.}
\label{tab:table1}
\begin{tabular}{
  >{\centering\arraybackslash}m{1.4cm}
  >{\centering\arraybackslash}m{2.6cm}
  >{\raggedright\arraybackslash}m{3.3cm}
}
\toprule
\textbf{Region} & \textbf{Electrodes} & \textbf{Core Functions} \\
\midrule
Frontal &
FP1, FPZ, FP2, AF3, AF4, F7, F5, F3, F1, FZ, F2, F4, F6, F8 &
Emotion regulation, decision-making, working memory, valence processing. \\
\midrule
Central &
FC3, FC1, FCZ, FC2, FC4, C3, C1, CZ, C2, C4 &
Somatosensory processing, motor control, sensorimotor integration. \\
\midrule
Parietal &
CP3, CP1, CPZ, CP2, CP4, P7, P5, P3, P1, PZ, P2, P4, P6, P8 &
Attention allocation, spatial perception, sensory integration. \\
\midrule
Left Temporal &
FT7, FC5, T7, C5, TP7, CP5 &
Language-related emotion, auditory processing, emotional memory. \\
\midrule
Right Temporal &
FC6, FT8, C6, T8, CP6, TP8 &
Non-verbal emotion, facial recognition, emotional prosody. \\
\midrule
Occipital &
PO7, PO5, PO3, POZ, PO4, PO6, PO8, CB1, O1, OZ, O2, CB2 &
Visual emotion processing, visual perception, scene analysis. \\
\bottomrule
\end{tabular}
\end{table}

After subject-specific alignment, the transformed feature $\mathbf{X} \in \mathbb{R}^{B \times T \times F}$ is obtained, where $B$ denotes the batch size and $T$ denotes the number of time steps. We then reshape $\mathbf{X}$ into a tensor $\mathbf{X}' \in \mathbb{R}^{B \times T \times N \times D}$ that explicitly encodes spatial and frequency-band information, where $N = 62$ denotes the number of electrodes and $D = 5$ denotes the number of frequency bands. For each time step $t$, a temporal slice $\mathbf{X}_t \in \mathbb{R}^{B \times N \times D}$ is extracted. Based on neuroanatomical principles and theories of hemispheric lateralization in emotion processing, we further partition the 62 electrodes into six mutually exclusive and collectively exhaustive functional brain regions:
\vspace{-0.5ex}
\begin{equation*}
\mathcal{R} =
\left\{
\begin{array}{lll}
\mathcal{R}_{\text{frontal}} &
\mathcal{R}_{\text{central}} &
\mathcal{R}_{\text{parietal}} \\[3pt]
\mathcal{R}_{\text{temporal}_l} &
\mathcal{R}_{\text{temporal}_r} &
\mathcal{R}_{\text{occipital}}
\end{array}
\right\}.
\end{equation*}
\vspace{-0.5ex}

The electrode index set for each region is denoted as 
$\mathcal{R}_r \subset \{0,1,\ldots,61\}$, satisfying 
$\bigcup_{r=1}^{6} \mathcal{R}_r = \{0,1,\ldots,61\}$ and 
$\mathcal{R}_i \cap \mathcal{R}_j = \varnothing \ (i \neq j)$.
This region partition serves as a structural inductive bias, confining attention interactions to electrodes within the same functional brain region and preventing non-physiological connections across functionally distinct regions. At each time step $t$, the query, key, and value representations are obtained via linear projections:
\begin{equation}
\mathbf{Q}_t = \mathbf{X}_t \mathbf{W}_q^{\top}, \quad
\mathbf{K}_t = \mathbf{X}_t \mathbf{W}_k^{\top}, \quad
\mathbf{V}_t = \mathbf{X}_t \mathbf{W}_v^{\top},
\end{equation}
where $\mathbf{W}_q$, $\mathbf{W}_k$, and $\mathbf{W}_v$ are learnable parameter matrices. The RGRM consists of two parallel branches designed to capture complementary interaction patterns within each brain region.

\textbf{Regional Continuous Branch (RCB)} based on the functional homogeneity assumption, electrodes within the same brain region tend to exhibit coordinated activation under specific emotional states. RCB extracts the overall activation pattern of each region by applying average pooling to the features of all electrodes belonging to that region, which enhances the stability and continuity of the learned representations. The computation can be expressed as:
\begin{equation}
\label{eq:rcb}
\mathbf{F}_{\mathrm{rcb}}^{(r)} =
\frac{1}{\lvert \mathcal{R}_r \rvert}
\sum_{i \in \mathcal{R}_r} \mathbf{V}_t^{(i)},
\end{equation}
where $\mathcal{R}_r$ denotes the electrode set of region $r$. This operation enhances the stability and continuity of regional representations.

\textbf{Regional Sparse Branch (RSB)} focuses on identifying the most discriminative sparse connections within a brain region. For each electrode $i$ in region $r$, its similarity with other electrodes in the same region is computed. The electrode exhibiting the strongest response is selected as the information source, excluding self-connections:
\begin{equation}
j = \arg\max_{k \in \mathcal{R}_r,\, k \neq i} 
\frac{\mathbf{Q}_t^{(i)} \cdot \mathbf{K}_t^{(k)}}{\sqrt{d_k}}, 
\quad 
\mathbf{F}_{\mathrm{rsb}}^{(i)} = \mathbf{V}_t^{(j)},
\end{equation}
where $d_k$ denotes the key dimension. This mechanism highlights critical intra-regional functional couplings while suppressing redundant information.
The outputs of the two branches are adaptively fused to obtain the final spatially enhanced features:
\begin{equation}
\mathbf{F}_{\mathrm{fused}}^{(i)} 
= \sigma(\alpha)\,\mathbf{F}_{\mathrm{rcb}}^{(i)} 
+ \bigl(1 - \sigma(\alpha)\bigr)\,\mathbf{F}_{\mathrm{rsb}}^{(i)},
\end{equation}
where $\alpha \in \mathbb{R}$ is a learnable scalar parameter and $\sigma(\cdot)$ denotes the sigmoid function. The fused features are aggregated via global average pooling and projected back to the original feature dimension:
\begin{equation}
\mathbf{F}_{\mathrm{pooled}} = \frac{1}{N} \sum_{i=1}^{N} \mathbf{F}_{\mathrm{fused}}^{(i)},
\end{equation}
\begin{equation}
\mathbf{F}_{\mathrm{out}} = \mathbf{F}_{\mathrm{pooled}} \mathbf{W}_o^{\top} + \mathbf{b}_o,
\end{equation}
where $\mathbf{W}_o$ and $\mathbf{b}_o$ denote the output projection parameters. 

Finally, a residual connection followed by layer normalization is applied:
\begin{equation}
\hat{\mathbf{X}}_t = \mathrm{LayerNorm}\bigl(\mathbf{X}_t + \mathbf{F}_{\mathrm{out}}\bigr).
\end{equation}
Based on the region-aware representations, explicit spatial attention weights are derived using a lightweight linear projection head with temporal aggregation:
\begin{equation}
\mathbf{a}_{\mathrm{spatial}}
=
\frac{1}{T}
\sum_{t=1}^{T}
\sigma\!\left(
\hat{\mathbf{X}}_t \mathbf{W}_{\mathrm{a}}^{\top}
+ \mathbf{b}_{\mathrm{a}}
\right),
\end{equation}
where $\mathbf{W}_{\mathrm{a}} \in \mathbb{R}^{F \times F}$ and $\mathbf{b}_{\mathrm{a}} \in \mathbb{R}^{F}$ denote the learnable parameters of the spatial attention head, and $\mathbf{a}_{\mathrm{spatial}} \in \mathbb{R}^{B \times F}$ represents the aggregated spatial attention weights. This design preserves the integrity of the original representations while incorporating region-level structural priors, enabling RGRM to encode neurophysiologically meaningful spatial dependencies for subsequent temporal modeling.

\subsection{Multi-Scale Temporal Transformer}
EEG-based emotional signals exhibit both short-term local dynamics and long-range dependencies across temporal scales, which are difficult to be jointly captured by a single temporal modeling mechanism. To this end, we propose MSTT module. Unlike the standard Transformer, which employs a global attention mechanism, MSTT explicitly models dependencies at different temporal scales by introducing structured attention masks. 
The module takes the RGRM-enhanced feature $\hat{\mathbf{X}} \in \mathbb{R}^{B \times T \times F}$ as input, and first projects it into a hidden space via linear transformation:
$\mathbf{H} = \hat{\mathbf{X}} \mathbf{W}_p^{\top}$,
where $\mathbf{W}_p \in \mathbb{R}^{H \times F}$ is the projection weight matrix and $H$ denotes the hidden dimension. 

At the core of MSTT are two parallel multi-head attention branches, each designed to capture temporal patterns at different scales. 
For the $h$-th attention head, the query, key, and value are computed as follows:
\begin{equation}
\mathbf{Q}^{(h)} = \mathbf{H} \mathbf{W}_Q^{(h)}, \quad
\mathbf{K}^{(h)} = \mathbf{H} \mathbf{W}_K^{(h)}, \quad
\mathbf{V}^{(h)} = \mathbf{H} \mathbf{W}_V^{(h)}.
\end{equation}

\textbf{The Local Temporal Dynamics Encoder} focuses on capturing local continuous patterns. 
This branch simulates a local receptive field by restricting each time point to attend only to other time points within its neighboring window:
\begin{equation}
\mathbf{M}_{\text{local}}[i,j] =
\begin{cases}
0 & |i-j| \le w \\
-\infty & \text{otherwise},
\end{cases}
\end{equation}
where $w$ denotes the size of the local window. 

The attention output under this constraint is given by:
\begin{equation}
\mathrm{Attn}_{\text{local}}^{(h)} =
\mathrm{Softmax}
\left(
\frac{\mathbf{Q}^{(h)} (\mathbf{K}^{(h)})^{\top}}{\sqrt{d_k}}
+ \mathbf{M}_{\text{local}}
\right)
\mathbf{V}^{(h)},
\end{equation}
Where $d_k$ denotes the feature dimension of a single attention head. 

The outputs from all attention heads are concatenated and linearly projected to obtain the local temporal representation:
\begin{equation}
\mathbf{H}_{\text{local}} =
\mathrm{Concat}
\left(
\mathrm{Attn}_{\text{local}}^{(1)}, \ldots,
\mathrm{Attn}_{\text{local}}^{(H_a)}
\right)
\mathbf{W}_o,
\end{equation}
where $\mathbf{W}_o \in \mathbb{R}^{H \times H}$ is the output projection matrix. 
This mechanism effectively captures short-term physiological dynamics in emotional responses.

\textbf{The Global Temporal Dependency Encoder} aims to establish long-range dependencies. To mitigate the computational cost associated with global attention, this branch employs a sparse sampling strategy, where each time point only connects to itself and periodically sampled key time points:
\begin{equation}
M_{\text{sparse}}[i,j] =
\begin{cases}
0, & \text{if } j = i \text{ or } j \equiv 0 \pmod{p}, \\
-\infty, & \text{otherwise}.
\end{cases}
\end{equation}
where $p$ denotes the sampling period. Under this mask constraint, the global attention for the $h$-th attention head is computed as:
\begin{equation}
\text{Attn}_{\text{global}}^{(h)} = \text{Softmax}\left( \frac{\mathbf{Q}^{(h)} (\mathbf{K}^{(h)})^\top}{\sqrt{d_k}} + M_{\text{sparse}} \right) \mathbf{V}^{(h)}.
\end{equation}

The outputs from all attention heads are concatenated and linearly projected to obtain the global temporal representation:
\begin{equation}
\mathbf{H}_{\text{global}} = \text{Concat}\left( \text{Attn}_{\text{global}}^{(1)}, \dots, \text{Attn}_{\text{global}}^{(H_a)} \right) \mathbf{W}_o.
\end{equation}

The outputs of the two temporal branches are first fused to integrate complementary local and global dynamics:
\begin{equation}
\mathbf{H}_{\mathrm{fused}} =
\phi\!\left(
\left[ \mathbf{H}_{\mathrm{local}};\, \mathbf{H}_{\mathrm{global}} \right]
\mathbf{W}_f^\top
\right),
\label{eq:fused}
\end{equation}
where $[\cdot;\cdot]$ denotes the concatenation operation, 
and $\mathbf{W}_f \in \mathbb{R}^{H \times 2H}$ is a learnable projection matrix. Based on the fused temporal representations, explicit temporal attention weights are computed and used to aggregate the sequence into a fixed-length representation:
\begin{equation}
\mathbf{a}_{\mathrm{temporal}}^{(t)}
=
\mathrm{Softmax}\!\left(
\mathbf{v}_a^\top
\tanh\!\left(
\mathbf{W}_a \mathbf{H}_{\mathrm{fused}}^{(t)}
\right)
\right),
\end{equation}
\begin{equation}
\mathbf{z}
=
\sum_{t=1}^{T}
\mathbf{a}_{\mathrm{temporal}}^{(t)}
\mathbf{H}_{\mathrm{fused}}^{(t)},
\end{equation}
where $\mathbf{a}_{\mathrm{temporal}} \in \mathbb{R}^{B \times T}$ denotes the normalized temporal attention distribution, and $\mathbf{z} \in \mathbb{R}^{B \times H}$ denotes the final multi-scale temporal representation.

\subsection{Collaboratively Optimized DG Strategy}
The core challenge in cross-subject EEG decoding lies in the significant distributional differences (domain shifts) observed among different subjects. To address this problem systematically, we propose a CoDG strategy. This method employs a multi-constraint joint optimization approach to learn domain-invariant feature representations, ensuring robust model performance on unseen subjects. Specifically, CoDG incorporates three complementary constraints: (i) Distribution Alignment Constraint – A mean-distance-based alignment loss (similar to Maximum Mean Discrepancy) that minimizes feature distribution shifts between subject groups by penalizing inter-subject mean discrepancies. (ii) Attention Consistency Constraint – An InfoNCE-style contrastive loss applied to spatial and temporal attention patterns, encouraging similarity in attention mechanisms within the same subject while maintaining cross-subject semantic consistency in spatiotemporal attention. (iii) Feature Structure Decoupling Constraint – A Frobenius norm penalty on the off-diagonal entries of the feature covariance matrix to reduce feature redundancy and promote structural disentanglement of representations. These constraints operate synergistically across three complementary dimensions: distribution, attention, and structure, thereby guiding the model toward learning emotion representations that are inherently more robust to individual variability.

The CoDG module takes the temporal representation $\mathbf{z} \in \mathbb{R}^{B \times H}$ output by the preceding MSTT module as input. It first maps the input features into a more generalizable subspace through a domain-invariant feature extractor:
\begin{equation}
\mathbf{f}_{\mathrm{inv}} = \phi_{\mathrm{inv}}(\mathbf{z}),
\end{equation}
where $\phi_{\mathrm{inv}}$ consists of a linear layer, layer normalization, and ReLU activation, designed to suppress subject-specific redundant variations.
To further enhance feature discriminability and decoupling, CoDG introduces a bias-free orthogonal projection matrix $\mathbf{W}_{\mathrm{orth}} \in \mathbb{R}^{H \times H}$ to transform the domain-invariant features:
\begin{equation}
\mathbf{f}_{\mathrm{orth}} = \mathbf{W}_{\mathrm{orth}} \mathbf{f}_{\mathrm{inv}}.
\end{equation}
This operation is optimized via an orthogonal constraint loss, aiming to maximize independence among different dimensions of the output features $\mathbf{f}_{\mathrm{orth}}$, thereby reducing redundancy and improving both interpretability and generalization capability.
To model the stable individual differences in spatiotemporal attention patterns, we encode the spatial and temporal attention weights into low-dimensional semantic embeddings:
\begin{equation}
\mathbf{e}_{\mathrm{spatial}}
= \phi_{\mathrm{spatial}}\!\left(\mathbf{a}_{\mathrm{spatial}}\right),
\end{equation}
\begin{equation}
\mathbf{e}_{\mathrm{temporal}}
= \phi_{\mathrm{temporal}}\!\left(\mathbf{a}_{\mathrm{temporal}}\right),
\end{equation}
where $\mathbf{a}_{\mathrm{spatial}} \in \mathbb{R}^{B \times F}$ and $\mathbf{a}_{\mathrm{temporal}} \in \mathbb{R}^{B \times T}$ denote the spatial and temporal attention weights derived from the RGRM and MSTT modules, respectively, and $\phi_{\mathrm{spatial}}$ and $\phi_{\mathrm{temporal}}$ are corresponding encoders.
These embeddings are then concatenated to form a unified attention representation:
\begin{equation}
\mathbf{e}_{\mathrm{attn}} = [\mathbf{e}_{\mathrm{spatial}}; \mathbf{e}_{\mathrm{temporal}}],
\end{equation}
which captures spatiotemporal emotion-related neural patterns and serves as the basis for subsequent contrastive learning.
To maximize the generalization capability of the learned features, the CoDG module employs the following three complementary loss functions as a joint optimization objective.
\subsubsection{Maximum Mean Discrepancy Loss (MMD Loss)}
To reduce feature distribution differences across subjects, a distributional constraint based on first-order statistical alignment is introduced. For any two distinct subjects $i \neq j$, the difference in their feature means is defined as:
\begin{equation}
\mathcal{L}_{\mathrm{mmd}} = \frac{1}{|P|} \sum_{(i,j) \in P} \left\| \mathbb{E}\left[ \mathbf{f}_{\mathrm{orth}}^{(i)} \right] - \mathbb{E}\left[ \mathbf{f}_{\mathrm{orth}}^{(j)} \right] \right\|_2,
\end{equation}
where $P = \{(i,j) \mid i < j,\, i,j \in S\}$ denotes the set of all distinct subject pairs within the current batch, $S$ is the set of unique subjects in the batch, and $\mathbf{f}_{\mathrm{orth}}^{(i)}$ represents the orthogonal feature subset for subject $i$.

\subsubsection{Contrastive Attention Loss}
Based on the InfoNCE framework, this loss encourages attention patterns from the same subject to be close while pushing apart those from different subjects:
\begin{equation}
\begin{aligned}
&\mathcal{L}_{\mathrm{contrast}}
=
-\frac{1}{B}
\sum_{i=1}^{B}
\log
\frac{
\sum_{j \neq i} \delta_{ij}
\exp\!\left(
\frac{\mathrm{sim}(\mathbf{e}_i,\mathbf{e}_j)}{\tau}
\right)
}{
\sum_{k \neq i} (1-\delta_{ik})
\exp\!\left(
\frac{\mathrm{sim}(\mathbf{e}_i,\mathbf{e}_k)}{\tau}
\right)
},
\end{aligned}
\end{equation}
where $\delta_{ij} = 1$ if and only if samples $i$ and $j$ originate from the same subject, $\mathrm{sim}(\cdot,\cdot)$ denotes cosine similarity, $\mathbf{e}_i$ denotes the attention embedding of the $i$-th sample, and $\tau$ is a temperature hyperparameter. This loss guides the model to learn attention-semantic representations that are consistent within individuals, thereby enhancing stability in cross-subject recognition.

\subsubsection{Orthogonal Constraint Loss}
To promote statistical independence among feature dimensions and reduce redundancy, we apply a decorrelation constraint to the batch features. First, the feature matrix is centered:
\begin{equation}
\widetilde{\mathbf{F}} = \mathbf{F} - \frac{1}{B} \mathbf{1} \mathbf{1}^\top \mathbf{F},
\end{equation}
where $\mathbf{F} \in \mathbb{R}^{B \times H}$ is the batch feature matrix and $\mathbf{1}$ is a column vector of ones. The correlation matrix is then computed as $\mathbf{R} = \frac{1}{B-1} \widetilde{\mathbf{F}}^\top \widetilde{\mathbf{F}}$, and the orthogonal loss is defined as:
\begin{equation}
\mathcal{L}_{\mathrm{orth}} = \left\| \mathbf{R} - \mathbf{I} \right\|_F^2,
\end{equation}
where $\mathbf{I}$ is the identity matrix and $\|\cdot\|_F$ denotes the Frobenius norm. This constraint penalizes off-diagonal elements, encouraging decorrelation across feature dimensions and thereby enhancing representation disentanglement and generalization capacity.

The final output of the CoDG module is the orthogonally projected feature representation $\mathbf{f}_{\mathrm{final}} = \mathbf{f}_{\mathrm{orth}}$, 
which serves as a domain-invariant emotional representation for subsequent classification.
These features are fed into a lightweight emotion classifier implemented as a multilayer perceptron (MLP) with batch normalization, ReLU activation, and dropout regularization, producing the predicted emotion probabilities via a log-softmax layer. For emotion recognition, the negative log-likelihood loss is employed as the classification objective:
\begin{equation}
\mathcal{L}_{\mathrm{cls}} = -\frac{1}{B} \sum_{i=1}^{B} 
\log \hat{\mathbf{y}}_{(i,c_i)},
\end{equation}
where $c_i$ denotes the true class index of the $i$-th sample, and $\hat{\mathbf{y}}_{(i,c_i)}$ represents the predicted log-probability of the corresponding class.
The overall training objective integrates the classification loss with the domain generalization constraints introduced in the CoDG module, forming a unified joint optimization framework:
\begin{equation}
\mathcal{L}_{\mathrm{total}} = \mathcal{L}_{\mathrm{cls}} 
+ \lambda_1 \mathcal{L}_{\mathrm{contrast}} 
+ \lambda_2 \mathcal{L}_{\mathrm{orth}} 
+ \lambda_3 \mathcal{L}_{\mathrm{mmd}},
\end{equation}
where $\lambda_1, \lambda_2, \lambda_3$ are hyperparameters that balance the contributions of the respective loss terms. Through this collaborative optimization strategy, the model jointly enhances emotion discriminability and cross-subject generalization capability.

\section{Experiments Setup}
\label{ex}
\subsection{Datasets and Preprocessing}
We evaluated RSM-CoDG on three public EEG datasets widely used in affective computing: SEED \cite{47}, SEED-IV \cite{48}, and SEED-V \cite{49}, all recording 62-channel signals via the international 10–20 system during video-induced emotion elicitation.

\textbf{SEED Dataset:} This dataset includes 15 subjects across three sessions, with each session containing 15 trials eliciting positive, negative, and neutral states.

\textbf{SEED-IV Dataset:} Comprising 15 subjects across three sessions, this dataset includes 24 trials per session corresponding to happiness, sadness, neutral, and fear.

\textbf{SEED-V Dataset:} This dataset involves 16 subjects across three sessions, where each session consists of 15 trials covering happiness, neutral, sadness, disgust, and fear.

\textbf{Preprocessing:} Following the standard processing pipeline consistent with the SEED dataset. The raw EEG signals were band-pass filtered within 0--75\,Hz and downsampled to 200\,Hz. Differential Entropy (DE) features \cite{50} were then extracted for experimental use. These DE features were computed across five frequency bands: $\delta$ (1--3\,Hz), $\theta$ (4--7\,Hz), $\alpha$ (8--13\,Hz), $\beta$ (14--30\,Hz), and $\gamma$ (31--50\,Hz). For each of the 62 EEG channels, DE values from all five bands were concatenated, resulting in an $F$-dimensional feature space with $F=310$ ($F = 62 \times 5$). To capture the temporal dynamics of EEG signals while preserving their time-frequency structure, we applied a sliding window approach to segment the DE features into sequential episodes. The window size was set according to the characteristics of each dataset: 30 time steps for SEED, 10 time steps for SEED-IV, and 18 time steps for SEED-V, ensuring compatibility with their respective experimental paradigms. This segmentation process generated a series of time-ordered episodes that retain the inherent time-frequency variations of the EEG signals. Finally, to mitigate amplitude variations across different channels and frequency bands, all features were normalized using min--max normalization.
\subsection{Experiment Settings and Evaluation Metrics}
Our model was trained using the Adam optimizer with an initial learning rate of $1 \times 10^{-4}$ and a weight decay of $5 \times 10^{-4}$. A StepLR scheduler (step size: 15 epochs, decay factor: 0.7) was applied to dynamically adjust the learning rate during training. To fully utilize multi-source domain samples in each training round, we adopted a two-level loop structure combining outer epochs and inner iterations. The batch sizes were set to 512 for the SEED dataset, 256 for SEED-IV, and 32 for SEED-V. The model was trained for a total of 120 epochs, with early stopping employed to prevent overfitting. Additionally, Gaussian noise with a standard deviation of 0.12 was injected during training to improve robustness. In terms of architecture, both RGRM and MSTT module used a hidden dimension of 64. Dropout with a rate of 0.4 was applied to the classifier and intermediate layers. All experiments were conducted using differential entropy features from single-session recordings. The implementation was carried out on an NVIDIA RTX 3090 GPU using the PyTorch framework, with a fixed random seed (seed=3) to ensure reproducibility.

To comprehensively evaluate the cross-subject generalization performance of the model, we adopted a Leave-One-Subject-Out (LOSO) cross-validation strategy. Specifically, in each LOSO iteration, one subject was designated as the target domain (test set), while the data from all remaining subjects constituted the source domain (training set). The process was repeated such that each subject served as the target domain exactly once, and the average test accuracy across all subjects was computed. The primary evaluation metric reported is the mean ± standard deviation of classification accuracy across all subjects. Additionally, complementary metrics—including F1-score, Sensitivity, and Specificity—were calculated to provide a more comprehensive assessment of classification performance.

\subsection{Comparative Methods}
To comprehensively evaluate RSM-CoDG, we benchmark it against representative approaches from three categories: traditional machine learning, DA and DG, and spatiotemporal representation learning.

\textbf{Traditional ML Baselines:} 
We include classical and widely-used classifiers, including Support Vector Machine (SVM) \cite{51}, K-Nearest Neighbors (KNN) \cite{52}, and Random Forest (RF) \cite{58}, as well as shallow transfer learning methods, including Geodesic Flow Kernel (GFK) \cite{53}, Subspace Alignment (SA) \cite{54}, Transfer Component Analysis (TCA) \cite{55}, and Correlation Alignment (CORAL) \cite{57}, which align feature distributions or latent subspaces across subjects.

\textbf{DA and DG Methods:} 
Representative methods include DAN \cite{59} (distribution alignment via MMD), DANN \cite{21} (adversarial domain-invariant learning), BiDANN \cite{45} (bi-hemisphere domain adversarial neural network), MS-MDA \cite{23} (multi-source DA with pairwise alignment), and DMMR \cite{25} (DG via noise injection and mixed mutual reconstruction).

\textbf{Spatiotemporal Representation Models:} 
These approaches capture EEG dynamics through specialized architectures,  including DGCNN \cite{10} (dynamic inter-channel graph learning), RGNN \cite{29} (graph modeling under biological topology priors), R2G-STNN \cite{30} (hierarchical region-to-global aggregation), EmT \cite{31} (temporal graph modeling with transformer-based context learning), AMDET \cite{41} (multi-dimensional spectral–spatial–temporal attention), and MAS-DGAT-Net \cite{60} (dynamic graph attention with multi-branch fusion).
\begin{table}[!t]
\centering
\footnotesize
\setlength{\tabcolsep}{1.5pt}

\begin{minipage}[t]{0.48\linewidth}
\centering
\caption{Performance comparison of RSM-CoDG with baselines on SEED Dataset.}
\label{tab:table2}
\begin{tabular}{p{1.9cm} c @{\hspace{4pt}} p{2.3cm} c}
\toprule
Methods & Acc$\pm$Std (\%) & Methods & Acc$\pm$Std (\%) \\
\midrule
\multicolumn{4}{c}{\textit{Traditional machine learning methods}} \\
\midrule
SVM \cite{51} & 56.73$\pm$16.29 & KNN \cite{52} & 55.26$\pm$12.43 \\
GFK \cite{53} & 56.71$\pm$12.29 & SA \cite{54} & 69.00$\pm$10.84 \\
TCA \cite{55} & 63.64$\pm$14.88 & KPCA \cite{56} & 61.28$\pm$14.62 \\
CORAL \cite{57} & 71.48$\pm$11.57 & RF \cite{58} & 62.78$\pm$06.60 \\
\midrule
\multicolumn{4}{c}{\textit{Deep learning methods}} \\
\midrule
DANN \cite{21} & 81.65$\pm$09.92 & BiDANN \cite{45} & 83.28$\pm$09.60 \\
DGCNN \cite{10} & 79.95$\pm$09.02 & RGNN \cite{29} & 79.00$\pm$14.80 \\
AMDET \cite{41} & 72.10$\pm$16.80 & MS-MDA \cite{23} & 81.43$\pm$10.17 \\
EmT-D \cite{31} & 80.20$\pm$11.50 & R2G-STNN \cite{30} & 84.16$\pm$07.63 \\
DMMR \cite{25} & 83.87$\pm$06.38 & MAS-DGAT \cite{60} & 80.02$\pm$05.79 \\
\midrule
\multicolumn{1}{l}{\textbf{RSM-CoDG}} & & & \textbf{86.35$\pm$07.17} \\
\bottomrule
\end{tabular}
\end{minipage}
\hfill
\begin{minipage}[t]{0.48\linewidth}
\centering
\caption{Performance comparison of RSM-CoDG with baselines on SEED-IV Dataset.}
\label{tab:table3}
\begin{tabular}{p{1.9cm} c @{\hspace{4pt}} p{2.3cm} c}
\toprule
Methods & Acc$\pm$Std (\%) & Methods & Acc$\pm$Std (\%) \\
\midrule
\multicolumn{4}{c}{\textit{Traditional machine learning methods}} \\
\midrule
SVM \cite{51} & 50.50$\pm$12.03 & KNN \cite{52} & 41.77$\pm$09.53 \\
GFK \cite{53} & 43.10$\pm$09.77 & SA \cite{54} & 34.74$\pm$05.29 \\
TCA \cite{55} & 44.11$\pm$10.76 & KPCA \cite{56} & 29.25$\pm$09.73 \\
CORAL \cite{57} & 48.14$\pm$10.38 & RF \cite{58} & 52.67$\pm$13.85 \\
\midrule
\multicolumn{4}{c}{\textit{Deep learning methods}} \\
\midrule
DAN \cite{59} & 59.27$\pm$14.45 & DANN \cite{21} & 57.16$\pm$12.61 \\
DGCNN \cite{10} & 52.82$\pm$09.23 & RGNN \cite{29} & 70.16$\pm$09.43 \\
BiDANN \cite{45} & 65.59$\pm$10.39 & MS-MDA \cite{23} & 61.43$\pm$15.71 \\
DMMR \cite{25} & 67.56$\pm$08.31 & MAS-DGAT \cite{60} & 70.22$\pm$09.12 \\
\midrule
\multicolumn{1}{l}{\textbf{RSM-CoDG}} & & & \textbf{71.59$\pm$09.78} \\
\bottomrule
\end{tabular}
\end{minipage}
\end{table}

\begin{table}[!htbp]
\centering
\footnotesize
\caption{Performance comparison of RSM-CoDG with competing baselines on SEED-V Dataset.}
\label{tab:table4}
\setlength{\tabcolsep}{1.5pt}
\begin{tabular}{p{2.1cm} c @{\hspace{5pt}} p{2.8cm} c}
\toprule
Methods & Acc$\pm$Std (\%) & Methods & Acc$\pm$Std (\%) \\
\midrule
\multicolumn{4}{c}{\textit{Traditional machine learning methods}} \\
\midrule
SVM \cite{51} & 53.14$\pm$10.10 & KNN \cite{52} & 35.73$\pm$07.98 \\
GFK \cite{53} & 38.32$\pm$10.11 & SA \cite{54} & 36.06$\pm$11.55 \\
TCA \cite{55} & 37.57$\pm$13.47 & KPCA \cite{56} & 35.47$\pm$09.39 \\
CORAL \cite{57} & 55.18$\pm$07.42 & RF \cite{58} & 42.29$\pm$16.02 \\
\midrule
\multicolumn{4}{c}{\textit{Deep learning methods}} \\
\midrule
DAN \cite{59} & 59.36$\pm$16.83 & DANN \cite{21} & 56.28$\pm$16.25 \\
DDC \cite{tzeng2014deep} & 55.45$\pm$17.15 & MS-MDA \cite{23} & 57.16$\pm$13.12 \\
DMMR \cite{25} & 60.31$\pm$11.47 & DCORAL \cite{61} & 56.26$\pm$14.56 \\
\midrule
\multicolumn{1}{l}{\textbf{RSM-CoDG}} & & & \textbf{62.77$\pm$08.86} \\
\bottomrule
\end{tabular}
\end{table}

\section{Results}
\label{raa}
\subsection{Overall Classification Results}
The evaluation results on the SEED dataset are presented in Table \ref{tab:table2}. Under the LOSO protocol, the proposed RSM-CoDG achieved a classification accuracy of 86.35\%, significantly outperforming all traditional machine learning methods. Among the traditional approaches, CORAL attained the highest accuracy of 71.48\%, while RSM-CoDG improved upon it by 14.87\%. Among deep learning methods, RSM-CoDG also achieved SOTA performance, surpassing the suboptimal model R2G-STNN (84.16\%) by 2.19\%. 

The results on the SEED-IV dataset are shown in Table \ref{tab:table3}. With the increase in the number of emotion categories to four, the overall task difficulty rises further. RSM-CoDG still achieved the best performance among both traditional and deep learning methods, with an accuracy of 71.59\%. Although the suboptimal model MAS-DGAT-Net enhances dynamic connection modeling through dynamic graph attention and multi-branch feature extraction, the absence of explicit cross-subject domain-invariant constraints renders it susceptible to individual variations under strict LOSO evaluation, limiting its accuracy to 70.22\%. 

To further evaluate the model's generalization capability and stability in multi-class scenarios, experiments were conducted on the SEED-V dataset with five emotion categories. The results are summarized in Table \ref{tab:table4}, which showed that RSM-CoDG achieved an accuracy of 62.77\%, significantly outperforming all comparative methods. RSM-CoDG achieves a 2.46\% improvement over the suboptimal DMMR, and further outperforms DA methods including DAN, DANN, and DCORAL by 3.41\%, 6.49\%, and 6.51\% respectively, as their exclusive reliance on distribution alignment or adversarial training aligns only specific statistical characteristics and remains insufficient for handling the complex individual variations in EEG signals.

Overall, RSM-CoDG demonstrates consistent superiority across all datasets.  It is noteworthy that the proposed method does not utilize any target domain data during training, whereas most baseline methods require access to such data for DA or parameter tuning. Despite this, RSM-CoDG achieves comparable or even superior cross-subject classification performance. These results indicate that the RSM-CoDG framework effectively disentangles emotion-related neural representations from confounding factors introduced by individual differences. By learning more generalizable emotion-related representations without relying on target domain information, the framework exhibits stronger adaptability and stability in real-world cross-subject scenarios. This provides robust support for its practical deployment on unseen subjects.
\begin{figure}[!t]
    \centering
    \begin{minipage}[t]{0.48\linewidth}
        \centering
        \includegraphics[width=\linewidth]{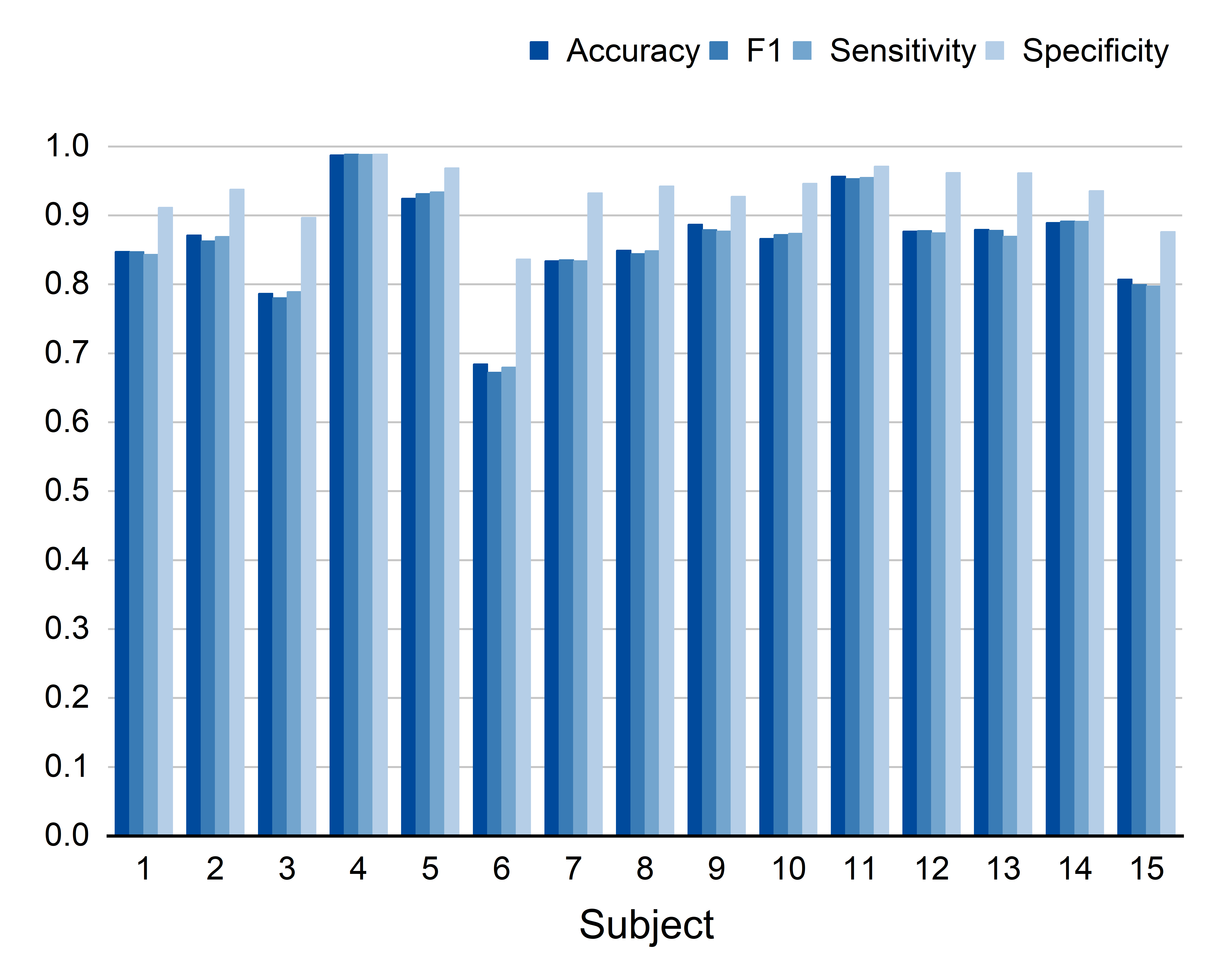}
        \caption{Cross-subject experimental results for every participant in the SEED dataset.}
        \label{fig:fig4}
    \end{minipage}
    \hfill
    \begin{minipage}[t]{0.48\linewidth}
        \centering
        \includegraphics[width=\linewidth]{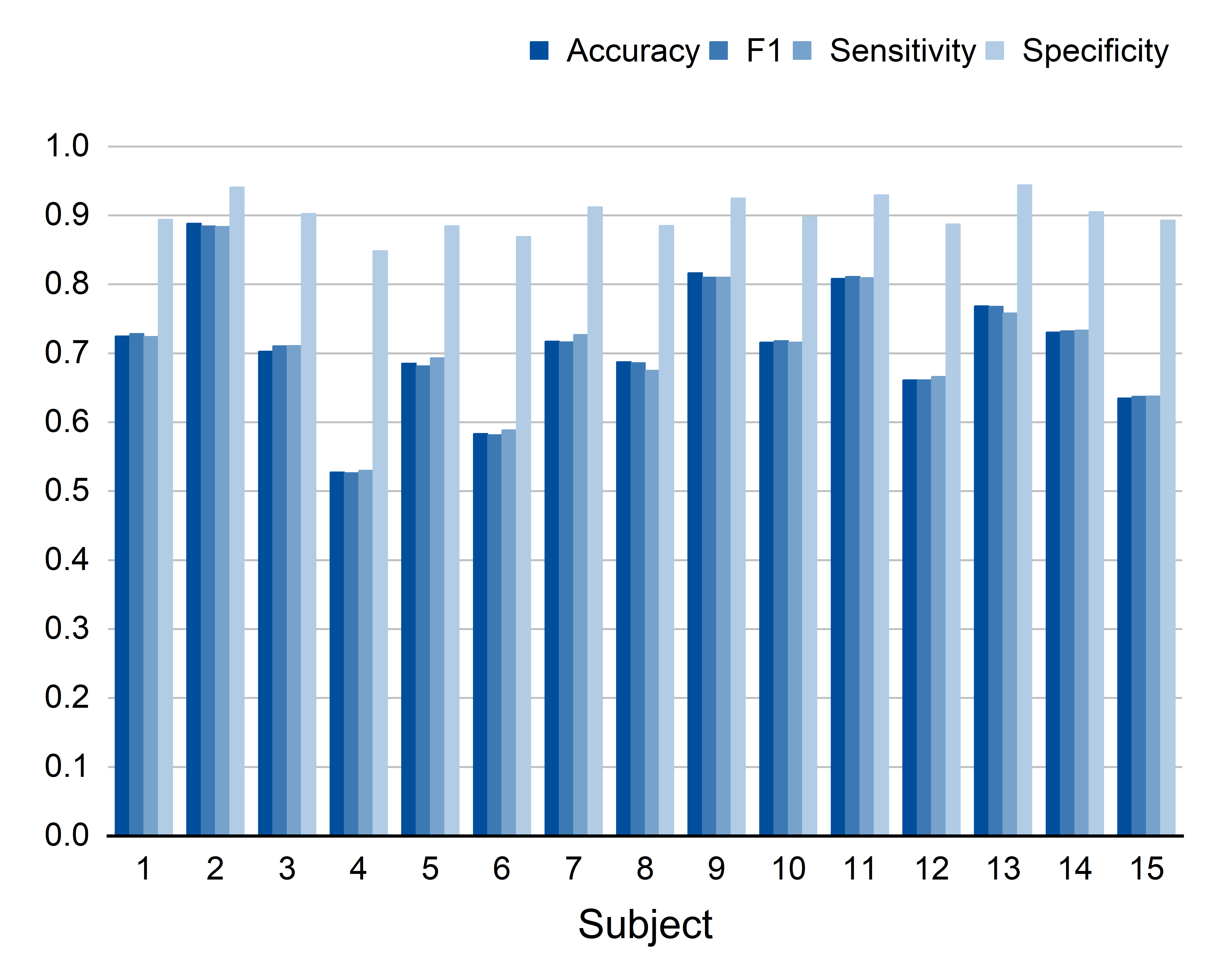}
        \caption{Cross-subject experimental results for every participant in the SEED-IV dataset.}
        \label{fig:fig5}
    \end{minipage}
\end{figure}

\begin{figure}[!t]
    \centering
    \includegraphics[width=0.48\linewidth]{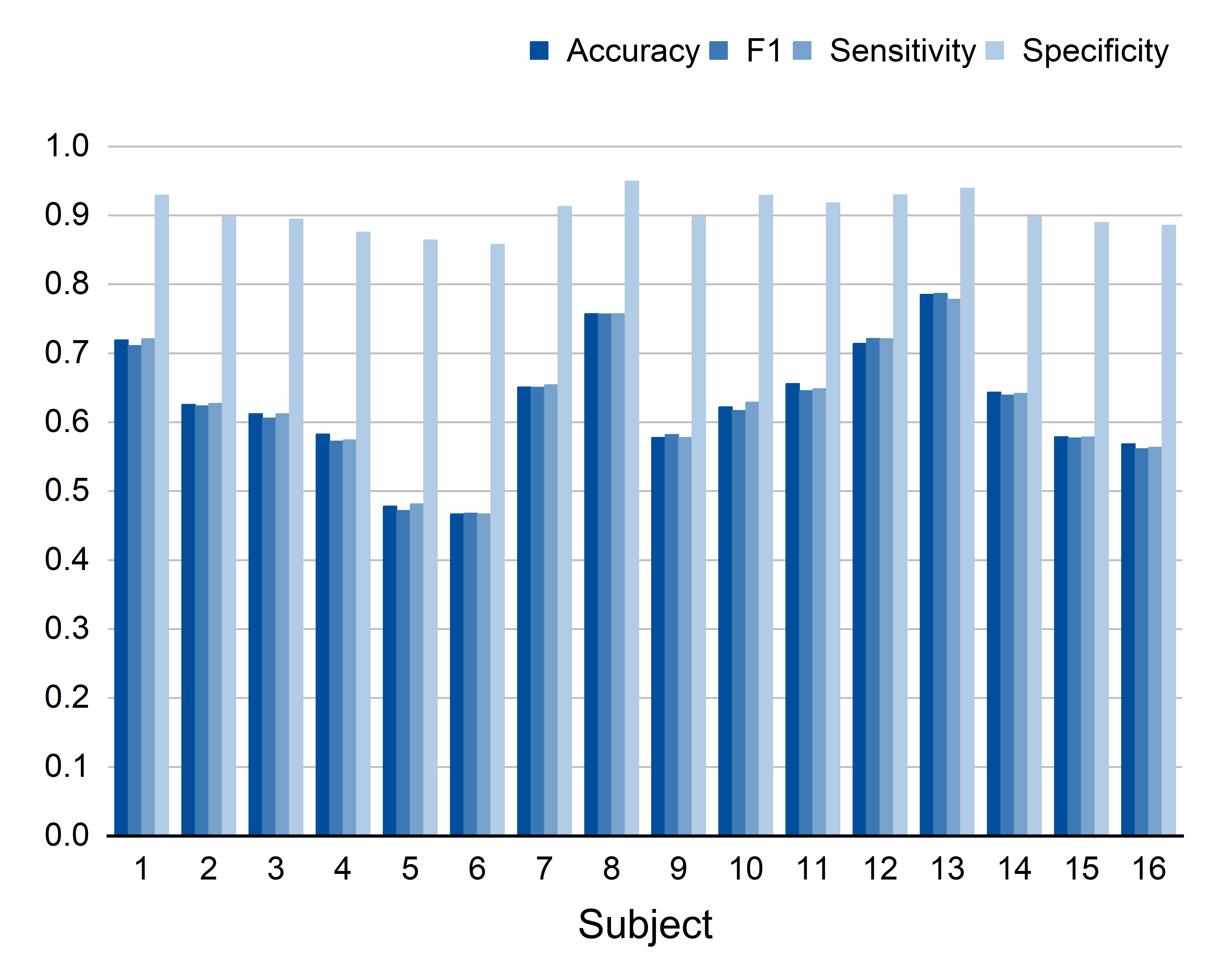}
    \caption{Cross-subject experimental results for every participant in the SEED-V dataset.}
    \label{fig:fig6}
\end{figure}

\subsection{Subject-Level Generalization Performance Evaluation}
Beyond overall accuracy, we evaluate individual performance for each subject using three complementary metrics: F1-score, sensitivity, and specificity. As shown in Fig.~\ref{fig:fig4}--\ref{fig:fig6}, the model maintains stable performance across the SEED (3-class), SEED-IV (4-class), and SEED-V (5-class) datasets, despite increasing task complexity. On the SEED dataset, RSM‑CoDG achieves average F1‑score, sensitivity, and specificity of 86.11\%, 86.18\%, and 93.32\%, with most subjects exceeding 80\% accuracy and low inter‑subject variation (std: ±7.17\%). On the four-class SEED-IV dataset, average metrics are 71.05\% (F1), 71.12\% (sensitivity), and 90.15\% (specificity), with most subjects above 60\% accuracy. On the more challenging SEED-V dataset, performance declines as expected but remains generalizable, reaching averages of 62.46\%, 62.73\%, and 90.45\%.

The observed performance trends and inter-subject variations can be attributed to several interrelated factors. Firstly, the gradual decline in accuracy from SEED to SEED‑V reflects the inherent increase in task difficulty with a greater number of emotion categories, leading to increased inter-class feature overlap and blurred decision boundaries. Secondly, inter-subject differences manifest as relatively weaker performance by a few subjects (e.g., subject 6 in SEED, subject 4 in SEED‑IV, subjects 5 and 6 in SEED‑V), which primarily stems from inherent physiological and psychological differences among individuals. Subjects with more distinct emotional expression features and differentiable physiological responses tend to achieve more reliable recognition (e.g., subject 4 in SEED, subject 2 in SEED‑IV, subjects 8 and 13 in SEED‑V), whereas those with subtler expressions or highly similar physiological response patterns exhibit relatively poorer classification results in the current feature space. Taken together, despite the progressive increase in task complexity, RSM-CoDG maintains stable cross-subject generalization, as evidenced by consistently low performance variance across datasets. This demonstrates the robustness of the proposed approach in multi-class EEG emotion recognition under substantial inter‑individual variability.

\subsection{Confusion Pattern Analysis}
\begin{figure*}
\centering
\includegraphics[width=\linewidth]{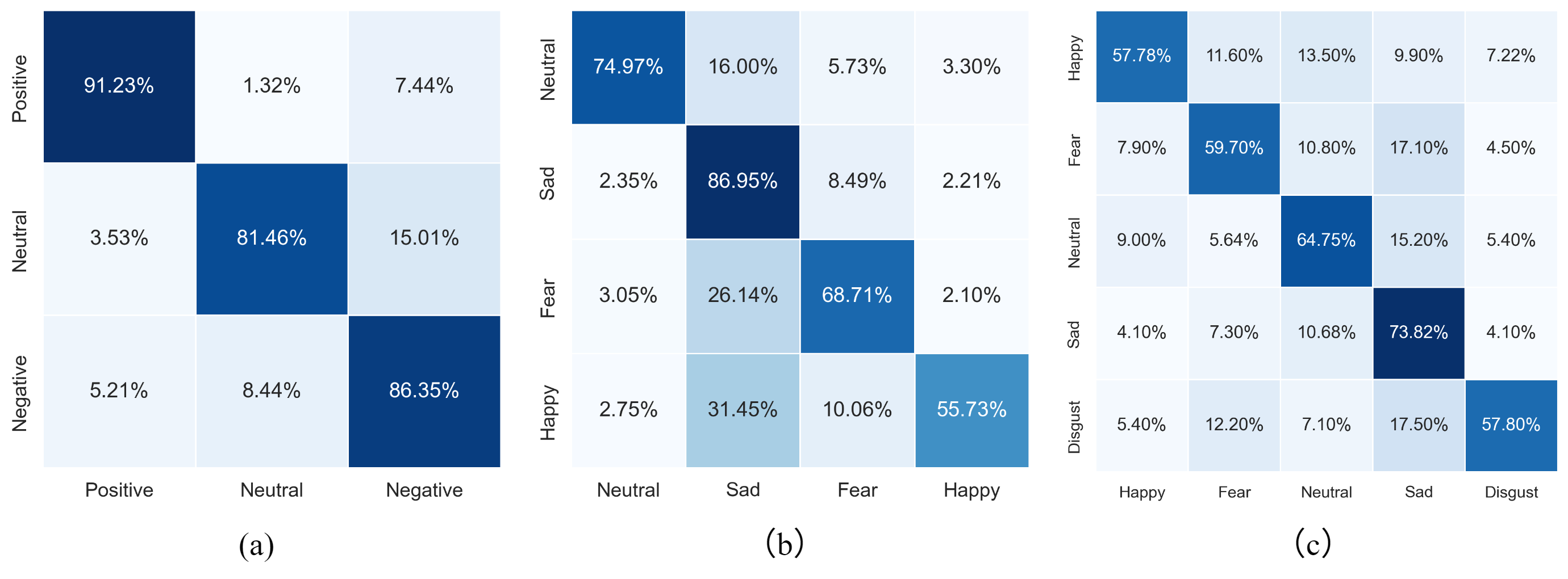}  
\caption{Confusion matrix for the SEED dataset (a), SEED-IV dataset (b), SEED-V dataset (c).}
\label{fig:fig7}
\end{figure*}

Fig. \ref{fig:fig7} presents the row-normalized confusion matrices for the SEED, SEED-IV, and SEED-V datasets, providing a detailed view of per-category classification performance. In each matrix, the vertical axis denotes the true emotion label, the horizontal axis represents the predicted category, and each cell $(i,j)$ indicates the proportion of samples from true class $i$ classified as class $j$. The color intensity along the diagonal corresponds to the classification accuracy for the respective emotion category. Across the three datasets, RSM-CoDG demonstrates stable and well-balanced cross-subject emotion recognition performance, with particularly strong discriminability for Sad and Positive emotions. On SEED, the model achieves the highest accuracy for Positive emotions, while moderate confusion is observed between Neutral and Negative classes, likely due to the weak affective cues of neutral EEG patterns. In the more challenging SEED-IV and SEED-V tasks, Sad consistently achieves the highest recognition accuracy, whereas Happy, Fear, and Disgust exhibit increased confusion, especially with Sad. This pattern suggests nonlinear overlaps in neural representations between high-arousal and negative-valence emotions, constraining classification performance in multi-class frameworks.

\subsection{Ablation Experiment}
\begin{table}[!t]
\centering
\setlength{\tabcolsep}{3pt} 
\caption{Ablation results of RSM-CoDG on the SEED, SEED-IV, and SEED-V datasets.}
\label{tab:table5}
\begin{tabular}{@{}>{\raggedright\arraybackslash}p{3cm}
                 >{\centering\arraybackslash}p{2.4cm}
                 >{\centering\arraybackslash}p{2.4cm}
                 >{\centering\arraybackslash}p{2.4cm}@{}}
\toprule
\multirow{2}{*}{\normalsize Model} & \multicolumn{3}{c}{\normalsize Accuracy $\pm$ Std (\%)} \\
\cmidrule(lr){2-4}
 & SEED & SEED-IV & SEED-V \\
\midrule
w/o SubjectAlign & 84.25$\pm$08.59 & 70.12$\pm$10.26 & 60.97$\pm$09.43 \\
w/o RGRM      & 82.32$\pm$07.34 & 67.57$\pm$09.38  & 58.34$\pm$10.86 \\
w/o MSTT      & 81.30$\pm$08.97 & 67.12$\pm$09.73  & 59.72$\pm$11.13 \\
w/o CoDG      & 80.10$\pm$09.62 & 65.12$\pm$11.65 & 57.63$\pm$10.27 \\
\textbf{full} & \textbf{86.35$\pm$07.17} & \textbf{71.59$\pm$09.78} & \textbf{62.77$\pm$08.86} \\
\bottomrule
\end{tabular}
\end{table}

\begin{figure}
\centering
\includegraphics[width=0.9\linewidth]{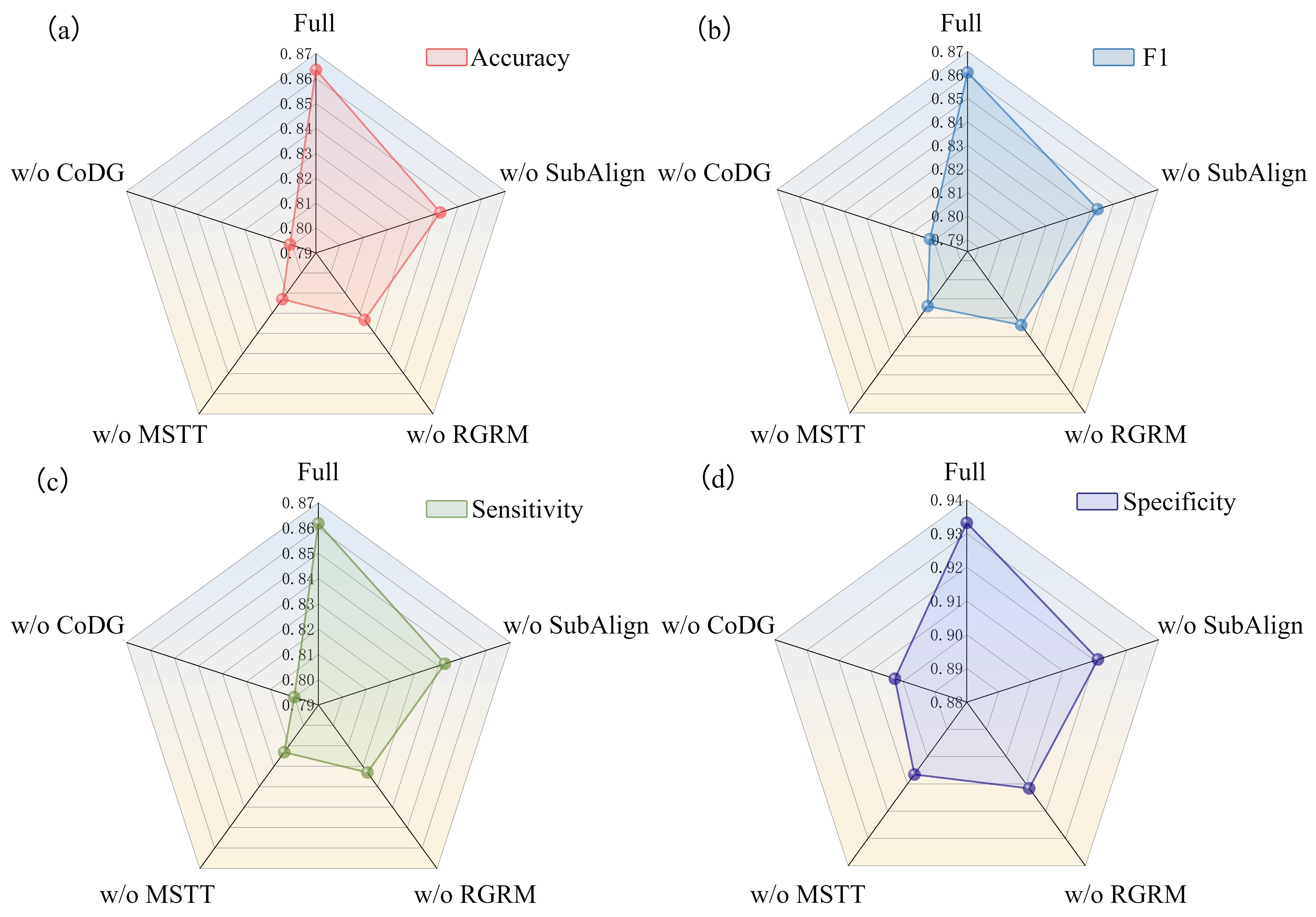}  
\caption{Detailed ablation study results of RSM-CoDG on the SEED dataset for cross-subject emotion recognition. Subfigures (a)–(d) report Accuracy, F1-score, Sensitivity, and Specificity, respectively, comparing the full model with different ablated variants.
}
\label{fig:fig8}
\end{figure}

\subsubsection{Component-Wise Analysis}
To validate the rationality and necessity of each module in RSM-CoDG, systematic ablation studies were conducted across the SEED, SEED-IV, and SEED-V datasets by sequentially removing key components. As shown in Table \ref{tab:table5}, the removal of any module led to a decline in model accuracy across all three datasets, preliminarily demonstrating the effectiveness of the overall architectural design. For a more comprehensive and intuitive illustration of each component's contribution, a radar chart (Fig. \ref{fig:fig8}) integrating four key metrics—Accuracy, F1-score, Sensitivity, and Specificity—is presented based on the SEED dataset. In all four subplots, the polyline representing the full model occupies the outermost position, achieving the highest values across every metric and significantly outperforming all ablated variants. Specifically: (1) Removing the Subject Alignment layer reduced the accuracy to 84.25\%, 70.12\%, and 60.97\% on SEED, SEED-IV, and SEED-V, respectively. This result indicates that lightweight subject-specific calibration plays a beneficial role in mitigating cross-subject individual variability. By learning a differentiable linear mapping for each subject, this layer significantly reduces distribution discrepancies at the input level, allowing RGRM and MSTT to focus on spatio-temporal pattern learning within a more consistent feature space. (2) Removing RGRM led to decreased accuracy of 82.32\%, 67.57\%, and 58.34\% on the three datasets, demonstrating that neuroscientific priors are crucial for spatial feature modeling. By incorporating functional brain region constraints, RGRM guides the structured organization of spatial information; its absence leads to a notable decline in the discriminative power of spatial representations. (3) Removing MSTT caused accuracy to drop to 81.30\%, 67.12\%, and 59.72\%, underscoring that effective modeling of emotion-related temporal dynamics is essential for accurate recognition. Without MSTT, the model struggles to capture both short-term fluctuations and long-range dependencies, thereby weakening its capacity to represent complex emotional dynamics. (4) Most critically, the removal of CoDG led to the most substantial performance decline with a drop of 6.25\%, 6.47\%, and 5.14\% across the three datasets. This result strongly demonstrates that relying solely on spatial or temporal modeling is insufficient to address individual variations in cross-subject scenarios. Without the multi-dimensional collaborative regularization provided by CoDG, the model struggles to effectively suppress inter-subject distribution discrepancies, thereby limiting its capacity to learn domain-invariant features.

\begin{table}[!t]
\centering
\setlength{\tabcolsep}{3pt} 
\caption{Ablation study of CoDG loss components on SEED, SEED-IV, and SEED-V datasets.}
\label{tab:table6}
\begin{tabular}{@{}>{\raggedright\arraybackslash}p{3.7cm}
                 >{\centering\arraybackslash}p{2.2cm}
                 >{\centering\arraybackslash}p{2.2cm}
                 >{\centering\arraybackslash}p{2.2cm}@{}}
\toprule
\multirow{2}{*}{\normalsize Loss} & \multicolumn{3}{c}{\normalsize Accuracy $\pm$ Std (\%)} \\
\cmidrule(lr){2-4}
 & SEED & SEED-IV & SEED-V \\
\midrule
w/o MMD Loss      & 82.95$\pm$09.89 & 68.57$\pm$11.06 & 59.37$\pm$10.16 \\
w/o Contrastive Loss  & 84.35$\pm$08.75 & 68.91$\pm$09.61 & 61.50$\pm$08.27 \\
w/o Orthogonal Loss   & 85.70$\pm$07.57 & 71.08$\pm$10.58 & 62.39$\pm$07.93 \\
\textbf{Full}        & \textbf{86.35$\pm$07.17} & \textbf{71.59$\pm$09.78} & \textbf{62.77$\pm$08.86} \\
\bottomrule
\end{tabular}
\end{table}
\subsubsection{CoDG Mechanism Validation}
To systematically evaluate the specific contributions of each constraint within the CoDG strategy, we conducted fine-grained ablation experiments on its three internally designed loss functions across the SEED, SEED-IV, and SEED-V datasets, as shown in Table~\ref{tab:table6}. The complete model (Full) achieved consistent performance gains, with accuracies of 86.35\%, 71.59\%, and 62.77\% on the three datasets, respectively. Removing the distribution alignment term (MMD) resulted in the most substantial degradation across all settings—accuracy drops to 82.95\% (–3.40\%) on SEED, 68.57\% (–3.02\%) on SEED-IV, and 59.37\% (–3.40\%) on SEED-V, demonstrating that explicit alignment of feature distributions is fundamental to learning domain-invariant representations, regardless of the number of emotion classes. Furthermore, ablating the attention consistency constraint (via contrastive learning) reduced accuracy to 84.35\% (–2.00\%), 68.91\% (–2.68\%), and 61.50\% (–1.27\%) on the three datasets, indicating that this constraint plays a key role in stabilizing the spatiotemporal patterns learned by RGRM and MSTT. In contrast, removing the feature orthogonality constraint led to a relatively smaller decrease to 85.70\% (–0.65\%) on SEED, 71.08\% (–0.51\%) on SEED-IV, and 62.39\% (–0.38\%) on SEED-V, suggesting that the orthogonality constraint provides an auxiliary yet consistent contribution to feature disentanglement across datasets. Overall, these results collectively validate that complementary constraints spanning distribution, attention mechanisms, and feature structure are indispensable for supporting generalization capability, and demonstrate that RSM-CoDG is a highly synergistic integrated framework whose performance stems from the coordinated interaction of its components, rather than from the isolated stacking of individual modules.

\subsubsection{Analysis of MSTT Attention Heads}
To systematically evaluate the impact of attention head quantity within the MSTT on feature representation, we conducted LOSO cross-validation on the SEED dataset using configurations of 4, 8 and 16 heads as details in Table \ref{tab:table7}. The results indicate that the configuration with eight attention heads achieves optimal classification accuracy (86.35\%) along with the fastest convergence (100 epochs), demonstrating that evenly dividing the hidden layer into subspaces enables effective capture of distinct temporal patterns. In contrast, the four-head configuration attains an accuracy of 83.71\%, where the limited number of heads may restrict the model’s ability to capture diverse temporal characteristics. Although the 16-head configuration offers greater parallelism, the reduced dimensionality per head compromises its representational capacity, leading to degraded performance (83.41\%) and slower convergence. In terms of convergence behavior, the eight-head model exhibits favorable stability and training efficiency, reaching performance saturation within an average of 100 epochs, outperforming both the four-head (120 epochs) and 16-head (150 epochs) configurations. This suggests that a moderate number of attention heads helps balance model complexity and training efficiency, providing adequate representational capacity while avoiding optimization challenges caused by structural redundancy. 

Furthermore, while the inference latency across all three configurations remains largely consistent at approximately 0.33 s per sample, the average training time per epoch is relatively high at around 130 s. This phenomenon is not primarily attributable to the multi-head attention mechanism itself but is rather a consequence of the CoDG strategy proposed in this study. During training, CoDG introduces multiple constraint-based loss functions, which require the computation of additional intermediate variables and gradients during both forward and backward propagation, thereby increasing overall training overhead. Nevertheless, under identical training conditions, the eight-head configuration demonstrates faster convergence and superior classification performance. This indicates that this configuration achieves a more favorable trade-off between representational capacity and optimization efficiency.
\begin{table*}[!t]
\centering
\renewcommand{\arraystretch}{1.2}
\caption{Performance comparison of RSM-CoDG with different numbers of attention heads in the MSTT module on the SEED dataset, including convergence speed, training efficiency, inference time, and classification accuracy.}
\label{tab:table7}
\begin{tabular}{
@{}
>{\centering\arraybackslash}p{1.9cm}
>{\centering\arraybackslash}p{3.0cm}
>{\centering\arraybackslash}p{3.0cm}
>{\centering\arraybackslash}p{3.0cm}
>{\centering\arraybackslash}p{3.0cm}
@{}
}
\toprule
\# Heads & Convergence Epoch & Avg Epoch Time (s) & Inference Time (s) & Acc$\pm$Std (\%) \\
\midrule
4  & 120 & 138.18 & 0.339 & $83.71 \pm 6.84$ \\
\textbf{8}  & \textbf{100} & \textbf{131.74} & \textbf{0.329} & $\mathbf{86.35 \pm 7.17}$ \\
16 & 150 & 134.38 & 0.326 & $83.41 \pm 8.45$ \\
\bottomrule
\end{tabular}
\end{table*}

\begin{figure}
\centering
\includegraphics[width=0.9\linewidth]{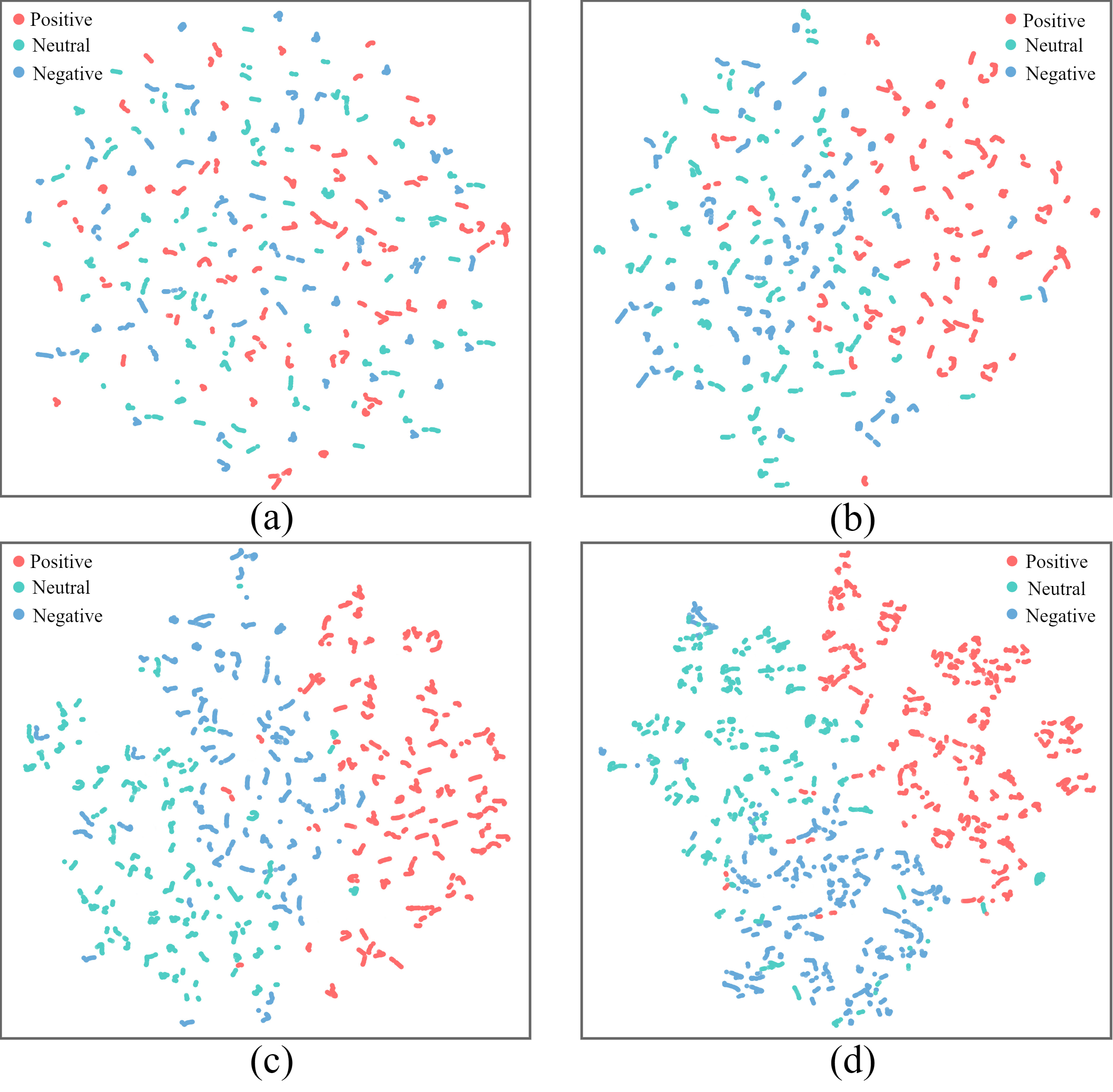}  
\caption{t-SNE visualization of feature representations at four processing stages: (a) original DE features, (b) features after RGRM, (c) features after MSTT, and (d) orthogonal features obtained by CoDG. Each point denotes one sample and is colored by its emotion label (red: positive, green: neutral, blue: negative).}
\label{fig:fig9}
\end{figure}
\subsection{Visualization and Interpretability Analysis}
\subsubsection{Feature Distribution Visualization}
To elucidate the progressive refinement of feature representations within the proposed architecture, we conducted t-SNE visualization analysis on the SEED dataset across four distinct processing stages. Specifically, we randomly selected 3,750 samples from 15 subjects and utilized PCA-based initialization to preserve the global structure, where red, green, and blue points represent positive, neutral, and negative emotions respectively. As illustrated in Fig.~\ref{fig:fig9}, the progressive evolution of emotion class separability becomes clearly observable across successive stages. In the original DE feature space (Fig.~\ref{fig:fig9}(a)), samples from the three emotion categories are heavily intermixed, exhibiting substantial overlap. After applying the RGRM (Fig.~\ref{fig:fig9}(b)), an initial separation among emotion clusters emerges. The spatial attention mechanism in RGRM effectively enhances region-specific activation patterns, leading to clearer class boundaries—particularly for positive emotions relative to neutral and negative ones. Nevertheless, considerable overlap remains between neutral and negative samples, indicating that spatial modeling alone is insufficient to disentangle closely related affective states. With the introduction of MSTT (Fig.~\ref{fig:fig9}(c)), the feature space is further refined. By explicitly modeling both short-term local dynamics and long-range temporal dependencies, MSTT substantially sharpens cluster boundaries and reduces intra-class dispersion. This improvement is especially evident in the diminished overlap between neutral and negative samples, underscoring the critical role of temporal dynamics in discriminating adjacent emotion levels. Finally, the CoDG strategy yields the most discriminative representations (Fig.~\ref{fig:fig9}(d)). Although a small degree of overlap between neutral and negative emotions persists, the overlap region is markedly reduced compared to earlier stages, and the three emotion classes become well separated.

These results quantitatively validate that RGRM, MSTT, and CoDG progressively contribute to extracting discriminative emotion-related features and effectively disentangling emotion-invariant components. The visualization outcomes align with the model's overall classification accuracy, jointly confirming that the proposed framework effectively captures emotion-specific patterns in EEG signals.
\begin{figure}
\centering
\includegraphics[width=0.9\linewidth]{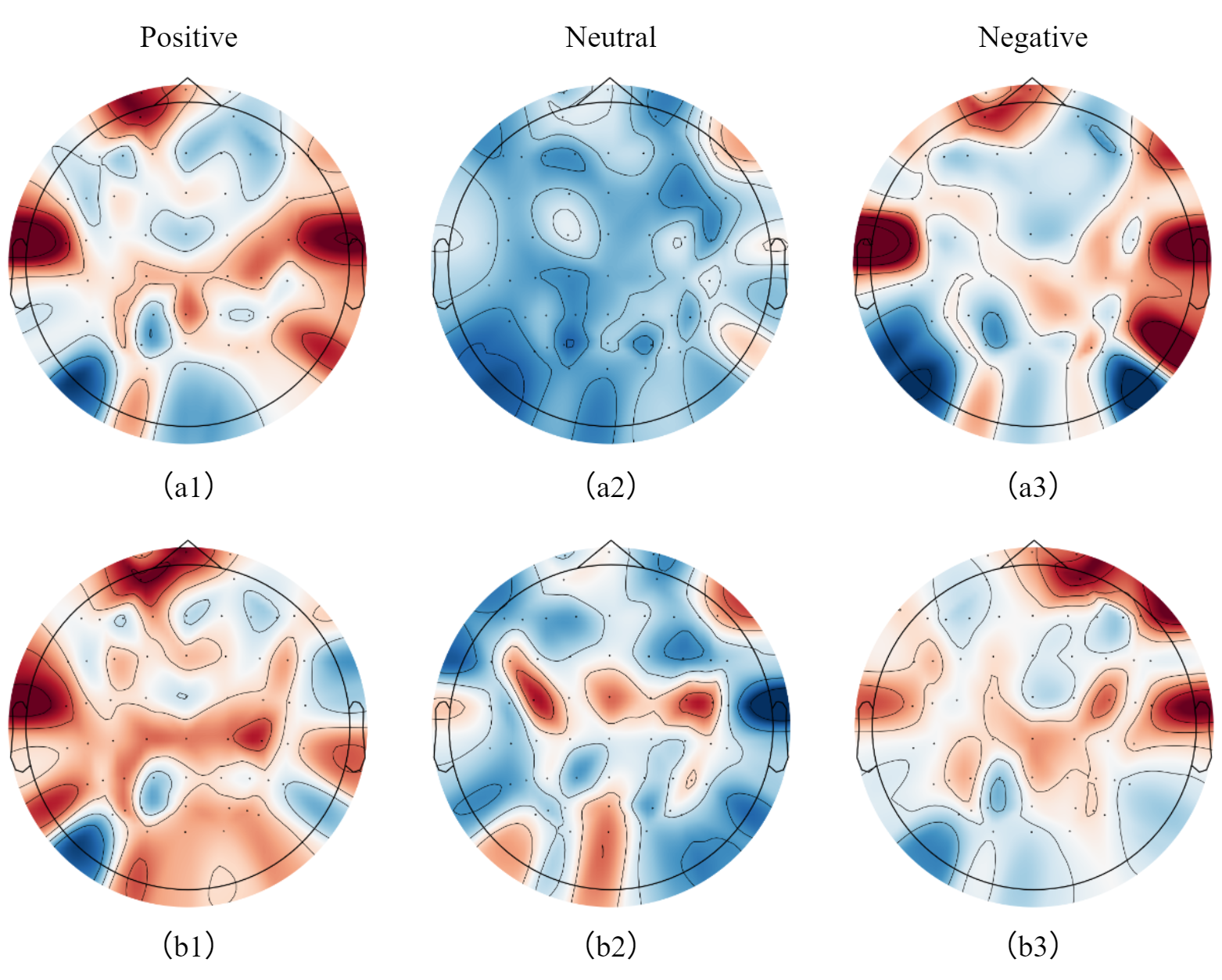}  
\caption{EEG activation maps for different emotional states. (a1) Subject 8 in a positive state, (a2) Subject 8 in a neutral state, (a3) Subject 8 in a negative state; (b1) Subject 3 in a positive state, (b2) Subject 3 in a neutral state, (b3) Subject 3 in a negative state.}
\label{fig:fig10}
\end{figure}

\subsubsection{Topographic Map Interpretability}
To investigate the interpretability of RSM-CoDG, we randomly selected two subjects from the SEED dataset and analyzed their EEG activity under three emotional states. For each emotional state, we visually mapped the feature contribution distributions of different brain regions onto a two-dimensional plane. This approach reveals how the model's internal spatial attention mechanism aligns with neurophysiological patterns, particularly in key functional areas such as the frontal, temporal, central, parietal, and occipital lobes.

Through the visualization of EEG topographic maps, we can observe the model's activation patterns across different emotional states, revealing the roles of multiple brain regions in emotional processing. For positive emotional states (Fig.~\ref{fig:fig10}(a1) and Fig.~\ref{fig:fig10}(b1)), the model displays the strongest activation in the frontal lobe—particularly the left frontal lobe—and the temporal lobe. The frontal lobe, which is typically associated with the induction of surprise and plays a central role in emotion regulation and control, while the temporal lobe is closely related to emotional experience and memory processing \cite{62,63,64}. Thus, the strong activation in these regions likely reflects the multidimensional processing underlying emotion regulation and affective experience during positive states. 
In negative emotional states (Fig.~\ref{fig:fig10}(a3) and Fig.~\ref{fig:fig10}(b3)), the model shows stronger responses in the frontal and temporal regions, particularly in the right hemisphere. This observation aligns with the emotional lateralization hypothesis, which posits that the right hemisphere is more involved in processing negative emotions \cite{65,66}. The heightened activation in the frontal lobe may be associated with the inhibition and regulation of negative emotions, while the co-activation of the temporal lobe supports the role of emotion–memory interactions in negative emotional generation. 
Under neutral emotional conditions (Fig.~\ref{fig:fig10}(a2) and Fig.~\ref{fig:fig10}(b2)), overall brain activation levels are relatively low. The model exhibits only mild spatial responses, with largely symmetrical activation between the left and right hemispheres. This pattern aligns with the neurophysiological characteristics of neutral emotional states, which are typically associated with lower emotional arousal and reduced cognitive load. 
In summary, our visualization analysis of EEG topographic maps demonstrates that the proposed RSM-CoDG model not only captures brain activation patterns consistent with established emotional neurobiological mechanisms but also reveals the coordinated involvement of multiple brain regions in emotional processing. This outcome enhances the transparency and interpretability of the model's decision-making process while providing robust evidence for its effectiveness and neurophysiological plausibility in cross-subject emotion recognition tasks.

\section{Conclusion}
\label{con}
This paper proposes RSM-CoDG, a Region-aware Spatiotemporal Modeling with Collaborative Domain Generalization framework for cross-subject EEG-based emotion recognition. Our approach innovatively leverages neuroscience-guided region-level spatial representations and multi-scale temporal dynamics modeling to guide feature extraction, while systematically addressing individual variability through multi-dimensional collaborative DG constraints in a unified architecture. The framework demonstrates how integrating neuroscientific priors with collaborative optimization can enhance emotion-related discriminative information and suppress non-emotional interference from cross-subject differences. Experimental results on the SEED, SEED-IV, and SEED-V datasets confirm that RSM-CoDG achieves stable and significant performance advantages in cross-subject recognition tasks, consistently outperforming mainstream approaches. However, the current implementation incurs notable computational overhead during training, primarily attributable to the increased complexity from joint optimization of multiple regularization constraints. Future work will therefore focus on more efficient implementations and lightweight strategies to reduce training costs and further improve inter-class recognition balance, facilitating the practical deployment of this paradigm in real-world affective computing.

\section*{Acknowledgments}
This work was supported by The Hong Kong Polytechnic University Start-up Fund (Project ID: P0053210), The Hong Kong Polytechnic University Faculty Reserve Fund (Project ID: P0053738), an internal grant from The Hong Kong Polytechnic University (Project ID: P0048377), The Hong Kong Polytechnic University Departmental Collaborative Research Fund (Project ID: P0056428), The Hong Kong Polytechnic University Collaborative Research with World-leading Research Groups Fund (Project ID: P0058097) and Research Grants Council Collaborative Research Fund (Project ID: P0049774).

\bibliographystyle{unsrt}  
\bibliography{references}

\end{document}